%% file: top.tex
\documentclass[10pt,twocolumn,letterpaper]{article}

\usepackage{cvpr}
\usepackage{times}
\usepackage{epsfig}
\usepackage{graphicx}
\usepackage{amsmath}
\usepackage{amssymb}
\usepackage{bm}
\usepackage{bbm}
\usepackage{algorithm}
\usepackage{algorithmic}
\usepackage{xparse,mathtools}
\usepackage{dsfont}
\usepackage[page]{appendix}
\usepackage{authblk}
\usepackage{float}
\usepackage[export]{adjustbox}
\usepackage{array}
\newcolumntype{?}{!{\vrule width 1pt}}
\newcolumntype{C}[1]{>{\centering\arraybackslash\hspace{0pt}}p{#1}}

\input{defs}

\usepackage[pagebackref=true,breaklinks=true,letterpaper=true,colorlinks,bookmarks=false]{hyperref}
\cvprfinalcopy 

\def\cvprPaperID{657} 

\ifcvprfinal\fi
\begin{document}

\title{Globally Consistent Multi-People Tracking using Motion Patterns}

\author[1]{Andrii Maksai}
\author[2]{Xinchao Wang}
\author[3]{Fran\c{c}ois Fleuret}
\author[1]{Pascal Fua}

\affil[1]{Computer Vision Laboratory, EPFL, Lausanne, Switzerland, \texttt{\small \{firstname.lastname\}@epfl.ch}}
\affil[2]{Beckman Institute, UIUC, Illinois, USA \texttt{\small \{firstname\}@illinois.edu}}
\affil[3]{IDIAP Reseach Institute, Martigny, Switzerland, \texttt{\small \{firstname.lastname\}@idiap.ch}}
%
\maketitle


\begin{abstract}

Many state-of-the-art  approaches to people  tracking rely on detecting  them in
each frame independently, grouping detections into short but reliable trajectory
segments, and then further grouping  them into full trajectories. This grouping
typically relies  on imposing local  smoothness constraints but almost  never on
enforcing more global constraints on the trajectories.

In this paper,  we propose an approach to imposing  global consistency by first
inferring  behavioral  patterns from the ground truth and  then  using  them  to guide  the  tracking
algorithm. \am{
When  used  in  conjunction  with  several  state-of-the-art  algorithms,  this
further increases  their already good  performance. Furthermore, we  propose an
unsupervised scheme  that yields almost  similar improvements without  the need
for ground truth.}

\end{abstract}

\vspace{-0.2cm}
\input{introduction}

\input{related}

\input{method}
\input{tracking}
\input{patterns}
\input{mot}

\input{results}
\input{conclusion}

{\small
  \bibliographystyle{ieee}
  \bibliography{string,vision,learning,optim,misc}
}

\onecolumn

\begin{appendices}
\input{supplementary}
\end{appendices}

\end{document}

%% file: defs.tex
\newcommand{\comment}[1]{}

\newif\ifdraft
\draftfalse

\ifdraft
 \newcommand{\PF}[1]{\textcolor{red}{\bf pf: #1}}
 \newcommand{\pf}[1]{{\color{red} #1}}
 \newcommand{\FF}[1]{\textcolor{green}{\bf ff: #1}}
 
 \newcommand{\AM}[1]{\textcolor{blue}{\bf am: #1}}
 \newcommand{\am}[1]{{\color{blue} #1}}
 \newcommand{\XW}[1]{\textcolor{cyan}{\bf xw: #1}}
 
\else
 \newcommand{\PF}[1]{}
 \newcommand{\pf}[1]{ #1 }
 \newcommand{\FF}[1]{}
 
 \newcommand{\AM}[1]{}
 \newcommand{\am}[1]{ #1 }
 \newcommand{\XW}[1]{}
 
\fi

\newcommand{\obj}[1]{\mathcal{#1}}
\newcommand{\argmin}{\operatornamewithlimits{arg\,min}}
\newcommand{\argmax}{\operatornamewithlimits{arg\,max}}

\newcommand{\MOTA}[0]{{\bf MOTA}}
\newcommand{\IDF} [0]{{$\text{\textbf{IDF}}_1$}}

\newcommand{\MDP}[0]{{\bf MDP}}
\newcommand{\SORT}[0]{{\bf SORT}}
\newcommand{\RNN}[0]{{\bf RNN}}
\newcommand{\KSP}[0]{{\bf KSP}}
\newcommand{\OUR}[0]{{\bf OUR}}
\newcommand{\Oracle}[0]{{\bf Oracle}}

\newcommand{\Towncentre}[0]{{\bf Town}}
\newcommand{\Station}[0]{{\bf Station}}
\newcommand{\ETH}[0]{{\bf ETH}}
\newcommand{\Hotel}[0]{{\bf Hotel}}
\newcommand{\PTZ}[1]{{\bf PTZ#1}}

%% file: introduction.tex

\section{Introduction}
\label{sec:introduction}

Multiple object  tracking (MOT) has a  long tradition  for applications such  as radar
tracking~\cite{Blackman86}. These early approaches gradually made their way into
vision community  for people  tracking purposes. They  initially relied  on Gating,
Kalman  Filtering ~\cite{BlackJ02,Mittal03b,Iwase04,Xu04,Magee04}  and later  on
Particle
Filtering~\cite{Giebel04,Smith05,Okuma04,Khan05,Yang05,Mauthner08,Breitenstein10}.
Because of their recursive nature, when  used to track people in crowded scenes,
they  are prone  to identity  switches and trajectory  fragmentations, which  are
difficult to recover from.

With the  recent improvements of  people detectors~\cite{Dollar12b,Benenson14},
the   Tracking-by-Detection   paradigm~\cite{Andriluka08}    has   now   become
the   preferred  way   to  solve   this  problem.   In  most   state-of-the-art
approaches~\cite{Tang15,Choi15,Milan16b,Xiang15},   this   involves   detecting
people in each frame independently, grouping detections into short but reliable
trajectory  segments (tracklets),  and then  further grouping  those into  full
trajectories.

While effective, existing tracklet-based approaches  tend to only impose local,
Markovian in  nature smoothness constraints  on the trajectories as  opposed to
more global  ones that stem from  people's behavioral patterns. For  example, a
person entering  a building  via a particular  door can be  expected to  head a
specific set of rooms  or a pedestrian emerging on the street  from a shop will
often turn left  or right to follow  the sidewalk. Such patterns  are of course
not  absolutes because  people  sometimes  do the  unexpected  but they  should
nevertheless inform the tracking algorithms. We know of  no existing technique
that imposes this  kind of global constraints in  globally optimal multi-target
tracking.

\input{figures/explanation}

Our first contribution is therefore an approach to first inferring patterns from
ground  truth data  and  then  using them  to  guide  the multi-target  tracking
algorithm.   More specifically,  we define  an objective  function that  relates
behavioral patterns to  assigned trajectories.  At training time,  we use ground
truth   data   to   learn   patterns   that  maximize   it,   as   depicted   by
Fig.~\ref{fig:explanation}(1,2).  At run time,  given these patterns, we connect
tracklets produced  by another algorithm, so  as to maximize the  same objective
function.   Fig.~\ref{fig:explanation}(3,4)   depicts  this  process.   We  will
demonstrate  that  when  used   in  conjunction  with  several  state-of-the-art
algorithms, this further increases their already good performance.

Our second contribution is to show we can obtain results almost as good
{\it without} ground-truth data  using an alternating scheme
that computes trajectories,  learns patterns from them, uses them  to compute new
trajectories, and iterates.

%% file: figures/explanation.tex
\begin{figure}[t]
\hspace{-1pt}
\hspace{-0.2cm}\includegraphics[trim={8cm 10.9cm 16.5cm 11.4cm},clip,width=0.5\textwidth]{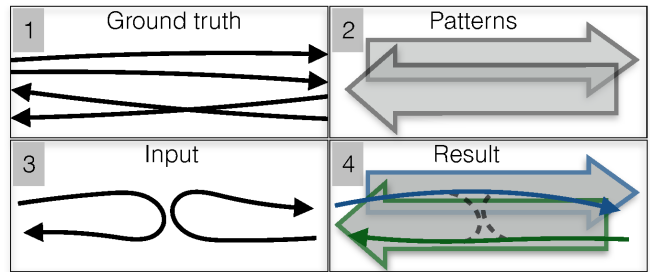}
\vspace{-0.4cm}
\caption{Given ground-truth  trajectories {\bf  (1)}, we learn  global patterns
{\bf (2)}. At  run-time, we start with trajectories found  by another algorithm
{\bf (3)}  to produce new  ones that are  consistent with the  learned patterns
{\bf (4)}.  Obtaining {\bf(1)} from  {\bf(2)} is  done at training  time, while
obtaining  {\bf(4)} from  {\bf(3)}  is  done during  testing.  If  there is  no
ground-truth,  we use  an iterative  scheme that  alternates between  computing
trajectories and learning patterns from them.}
\label{fig:explanation}
\vspace{-0.5cm}
\end{figure}

%% file: related.tex
\section{Related Work}
\label{sec:related}

\comment{Multiple  Object Tracking  (MOT), scene  modeling, pattern  mining, and  anomaly
detection all have a long history  in the computer vision community.}

We    briefly     review    data    association    and     behavior    modeling
techniques.~\cite{Wang13b,Li13b} contain more complete  overview of the topics.
We also discuss the metrics for MOT evaluation.

\subsection{MOT as Data Association}

Finding the right trajectories linking the detections, or data association, has
been formalized  using various models. For real-time  performance, data
association often relies either on matching locally between existing tracks and
new targets~\cite{Fagot-Bouquet16,Lenz15,Bae14,Choi15,Milan16}  or on filtering
techniques~\cite{Oh09,Rodriguez11}. The resulting  implementations are fast but
often perform less  well than batch optimization methods, which  use a sequence
of frames to  associate the data optimally  over a whole set  of frames, rather
than greedily in each next frame.

Batch optimization  is usually  formulated as a
shortest  path problem~\cite{Berclaz11,   Pirsiavash11},  network  flow problem~\cite{Zhang08a},
generic  linear  programming~\cite{Jiang07}, integer  or  quadratic
programming~\cite{Leibe07b,Brendel11,Wang14e,Hamid15,Dehghan15b,Yu16}.
A  common way  to reduce  the computational  burden is  to first  group reliable
detections  into short  trajectory fragments  known as  tracklets and  then reason on
these            tracklets            instead           of            individual
detections~\cite{Joo07,Singh08,Li09,Kuo11,BenShitrit14}.

However,  whether or  not tracklets  are used,  making the  optimization problem
tractable  when looking  for  a global  optimum limits  the  class of  objective
functions that can be used.  They are usually restricted to functions that can be
defined on  edges or  edge pairs in  a graph whose  nodes are  either individual
detections or tracklets. In other words, such objective functions can be used to
impose relatively local  constraints.
To impose global constraints, the objective functions have to involve
multiple people and long time spans.
They are solved using  gradient descent with exploratory  jumps~\cite{Milan14}, inference
with a dynamic graphical model~\cite{Choi15},  or iterative groupings of shorter
tracklets                               into                              longer
trajectories~\cite{Kuo10,Fragkiadaki12,Andriyenko12}. However, this comes at the
cost of losing any guarantee of global optimality.

By  contrast, our  approach is  designed for  batch optimization  and finding  the
global optimum, while using an objective function that is rich enough to express
the relation between global trajectories and non-linear motion patterns.

\subsection{Using Behavioral Models}

There have been a number of attempts  at incorporating human behavioral models into
tracking algorithms  to increase their  reliability.  For example,  the approaches
of~\cite{Pellegrini10,Alahi16} model collision avoidance behavior to improve
tracking, the  one of~\cite{Yi16c}  uses behavioral model  to predict  near future
target locations, and the one  of~\cite{Qin12} encodes local velocities into the
affinity  matrix of  tracklets.   These approaches  boost the performance but  only
account for very local interactions,  instead of global behaviors that influence
the {\it whole} trajectory.

Many  approaches  to  inferring  various forms of global  patterns
have  been  proposed over  the  years~\cite{Ravanbakhsh16,Kalayeh15,Mahadevan10,Piciarelli05,Zelniker08,Hu04,Calderara11,Kim09,Li14c}.
However,       the       approaches       of~\cite{Berclaz08b},~\cite{Kratz12},
and~\cite{Basharat08} are  the only ones we  know of that attempt  to use these
global  patterns to  guide  the tracking.  The  method of~\cite{Berclaz08b}  is
predicated  on the  idea that  behavioral maps  describing a  distribution over
possible  individual movements  can be  learned and  plugged into  the tracking
algorithm to  improve it. However,  even though the  maps are global,  they are
only  used  to  constrain  the  motion  locally  without  enforcing  behavioral
consistency  over  the  whole   trajectory.  In~\cite{Basharat08},  an  E-M-based
algorithm  is used  to model  the scene  as a  Gaussian mixture  that
represents the  expected size  and speed  of an object  at any  given location.
While  the  model can  detect  global  movement  anomalies and  improve  object
detection, the motion  pattern information is not used to  improve the tracking
explicitly.  In~\cite{Kratz12},  modeling  of  the optical  flow  improves  the
tracking, and helps to detect anomalies, but it relies on the presence of dense
crowds, motion flow of which is used for tracking.

\comment{it is based solely  on individual speed predictions at every
point  along the  trajectory,  which  is not  appropriate  for truly  non-linear
motions.  \XW{I don't understand the last sentence.} \FF{Is that that the speed
is not constrained,  and may not be constant?} \AM{I  think last sentence might
not be  entirely true. Would it  be better to  rephrase saying that it  is "not
appropriate when there are multiple overlapping global movement patterns?"}}

\comment{
Notable differences are that our approach uses clearly defined patterns with the
direction of  movement, while  maps can  contain pieces  of several  patterns or
loops. Additionally, our approach is able  to detect anomalies which are global,
rather than local,  in the observed trajectories, and requires  a minimum amount
of  training data.
}

\subsection{Quantifying Identity Switches}
\label{sec:metrics}

\input{figures/metrics}

\input{figures/graph}

In this  paper, we aim to do globally consistent tracking by preventing  {\it identity switches}  along reconstructed
trajectories,  for example when  trajectories of  different people  are
merged into  one or when a  single trajectory is  fragmented into  many.  We
therefore need an appropriate metric to gauge the performance of our algorithms.

The set  of CLEAR  MOT metrics~\cite{Bernardin08} has  become a  {\it de-facto}
standard for evaluating tracking results. Among these, Multiple Object Tracking
Accuracy  (MOTA) is  the  one that  is  used most  often  to compare  competing
approaches.  However, it  has  been pointed  out that  MOTA  does not  properly
account for identity  switches~\cite{BenShitrit11,Yu16,Bento16}, as depicted on
the left  side of Fig.~\ref{fig:metrics}.  More adapted metrics  have therefore
been  proposed. For  example, \IDF{}  is computed  by matching  trajectories to
ground-truth so as  to minimize the sum of  discrepancies between corresponding
ones~\cite{Ristani16}.  Unlike \MOTA{},  it  penalizes switches  over the  {\it
whole}  trajectory  fragments  assigned  to the  wrong  identity,  as  depicted
by  the  right side  of  Fig.~\ref{fig:metrics}.  Furthermore, unlike  Id-Aware
metrics~\cite{Yu16,BenShitrit11}, it does not require knowing the true identity
of the people being tracked, making it more widely applicable.

\comment{A metric very
similar  to  \IDF{}  was  proposed in~\cite{Bento16}.  It  features  additional
emphasis on  rigorous mathematical definition and  provides additional evidence
of interest in going beyond \MOTA{}.}

In the  results section,  we report  our results in terms  of both \MOTA{}
because  it is  widely used  and \IDF{} to  highlight the  drop in  identity
switches our method brings about.

%% file: figures/metrics.tex

\begin{figure}[t]
\begin{center}
\hspace{-1pt}
\begin{tabular}{cc}
  \multicolumn{2}{c}{
  \hspace{-0.2cm}\includegraphics[trim={0.16cm 1.0cm 0.66cm 2.66cm},clip,width=0.48\textwidth]{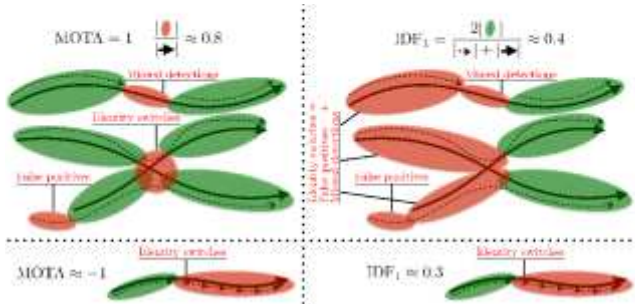}} 
\end{tabular}
\end{center}
\vspace{-0.4cm}
\caption{Effect of  identity switches on  the tracking metrics. The thick
lines represent  ground-truth trajectories and  the thin dotted  ones recovered
trajectories. The trajectory fragments that count positively are shown in green
and those that count  negatively in red. The formulas at the  top of the figure
depict graphically how  the \MOTA{} and \IDF{} scores are  computed. {\bf Top}:
Three ground-truth  trajectories, with the  bottom two crossing in  the middle.
The four recovered  trajectories feature an identity switch where  the two real
trajectories intersect, missed detections  resulting in a fragmented trajectory
and therefore another  identity switch at the top, and  false detections at the
bottom left. When using \MOTA{}, the identity switches incur a penalty but only
very  locally,  resulting in  a  relatively  high  score. By  contrast,  \IDF{}
penalizes the recovered  trajectories over the {\it  whole} trajectory fragment
assigned to the wrong identity, resulting  in a much lower score. {\bf Bottom:}
The last two thirds of the  recovered trajectory are fragmented into individual
detections that are not linked. \MOTA{}  counts each one as an identity switch,
resulting in a  negative score, while \IDF{} reports a  more intuitive value of
0.3.}
\label{fig:metrics}
\vspace{-0.5cm}
\end{figure}

%% file: figures/graph.tex

\begin{figure*}[t!]
\begin{center}
\hspace{-1pt}
\begin{tabular}{ccc}
  \hspace{-0.2cm}\includegraphics[trim={9cm 11cm 10.5cm 5cm},clip,width=0.31\textwidth]{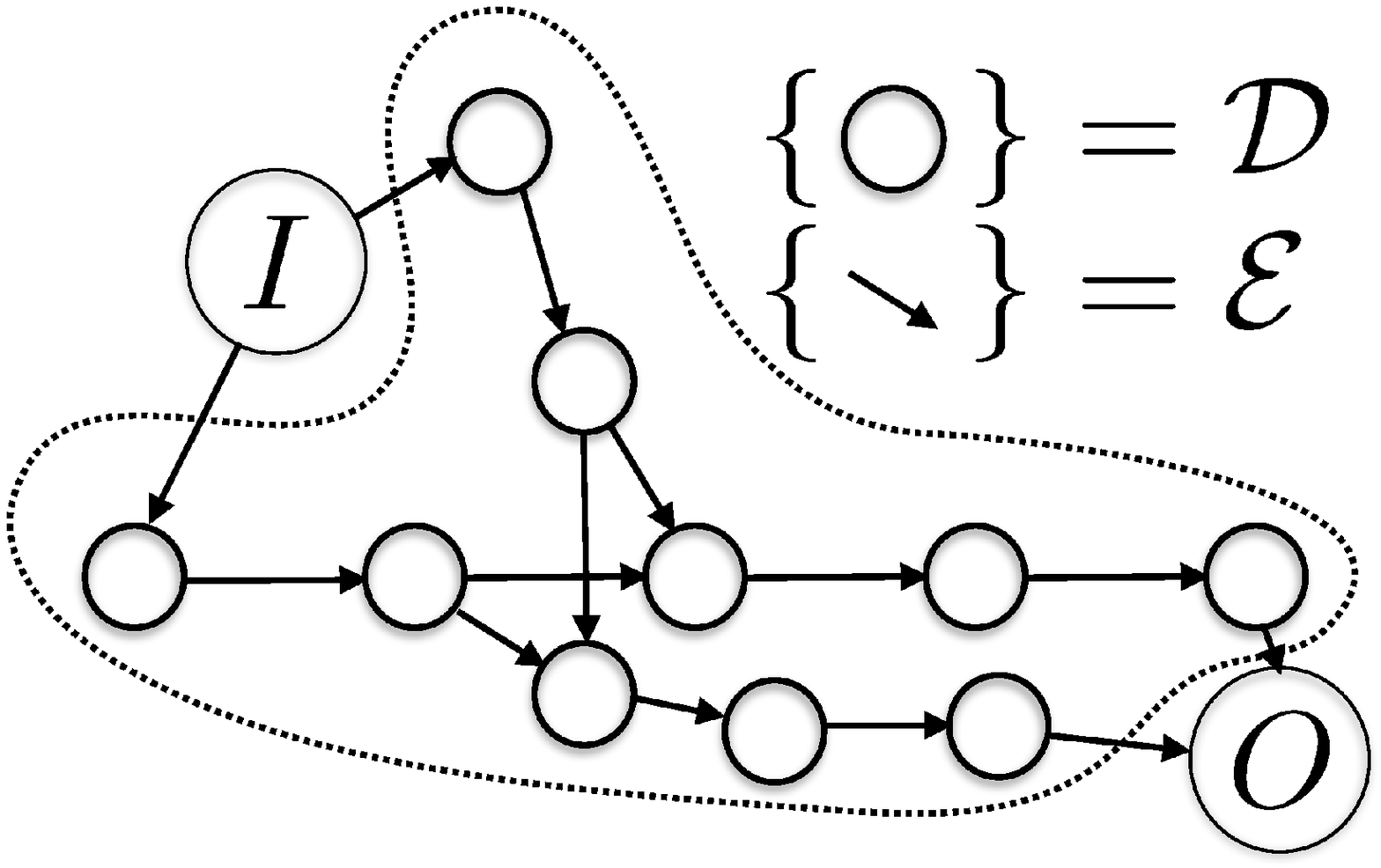} &
  \hspace{-0.2cm}\includegraphics[trim={9cm 11cm 9cm 5cm},clip,width=0.345\textwidth]{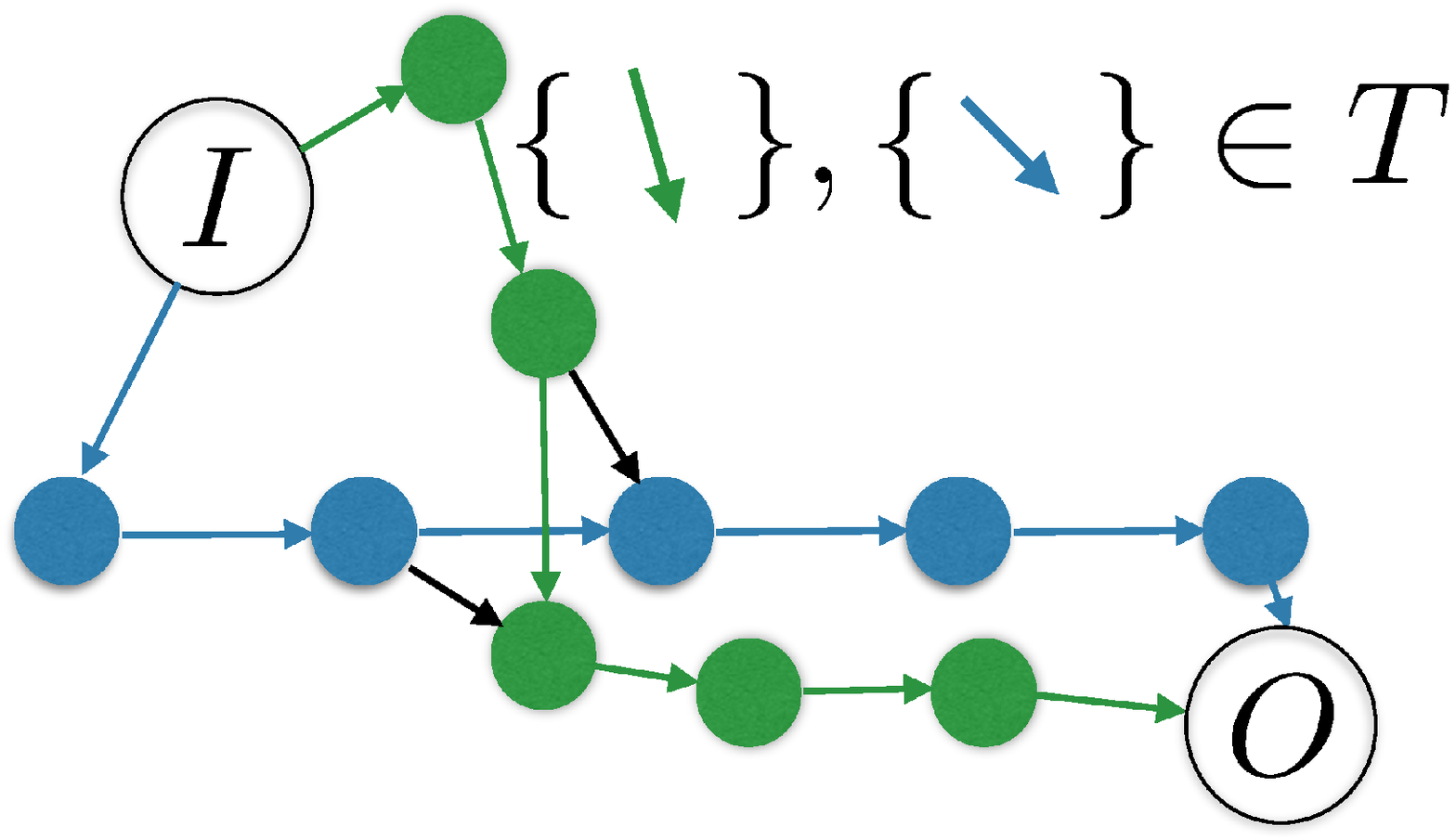} &
  \hspace{-0.2cm}\includegraphics[trim={4.5cm 5.5cm 5cm 2.5cm},clip,width=0.325\textwidth]{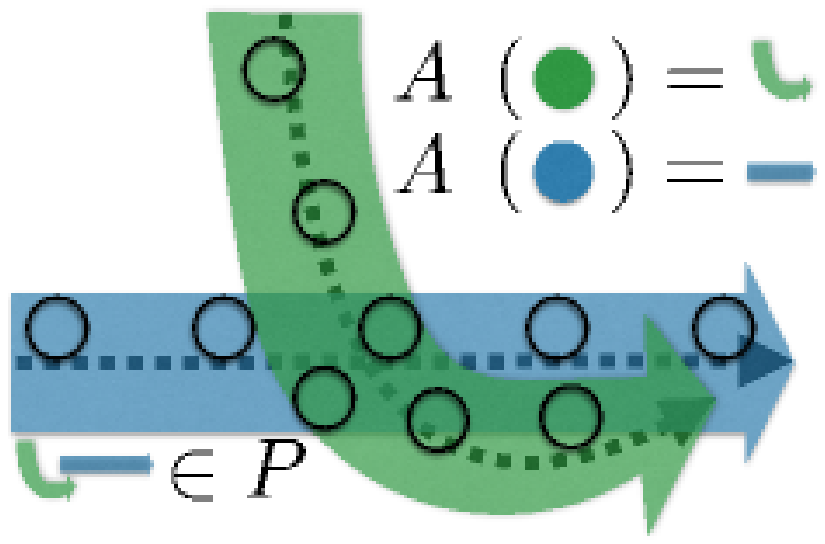}
\\
   \hspace{-0.2cm} (a) &
   \hspace{-0.2cm} (b) &
   \hspace{-0.2cm} (c)
\end{tabular}
\end{center}
\vspace{-0.4cm}
\caption{\textbf{(a)} Given a set  of high-confidence detections $\obj{D}$, and
a  set  of  allowed  transitions  $\obj{E}$,  we  seek  to  find:  \textbf{(b)}
trajectories of the people, represented by transitions from $T$; \textbf{(c)}
a set  of behavioural patterns $P$,  which define where people  behaving in a
particular way are  likely to be found; an assignment  $A$ of each individual
detection  to a  pattern,  specifying  which pattern  did  the  person in  this
detection follow.}
\label{fig:graph}
\vspace{-0.5cm}
\end{figure*}

%% file: method.tex
\section{Formulation}
\label{sec:Formulation}

In this  section, we formalize the  problem of discovering and  using behavioral
patterns to impose global constraints  on a multi-people tracking algorithm. In
the  following sections  we  will  use it  to  estimate  trajectories given  the
patterns and to discover the patterns given ground-truth trajectories.

\subsection{Detection Graph}
\label{sec:graph}

Given  a set  of  high-confidence detections  $\obj{D} =  \{1,  \dots, L\}$  in
consecutive  images of  a  video sequence,  let $\obj{V}=\obj{D}\cup\{I,  O\}$,
where $I$ and $O$ denote possible trajectory start and end points and each node
$v  \in \obj{D}$  is  associated  with a  set  of  features \comment{$f(v)  \in
\obj{F}$} that encode location, appearance,  or other important properties of a
detection. Let $\obj{E}  \subset \obj{V}^2$ be the set  of possible transitions
between the detections.  $\obj{G}=(\obj{V}, \obj{E})$ can then be  treated as a
{\it detection graph} of which the desired trajectories are subgraphs. As shown
by Fig.~\ref{fig:graph}, let
\begin{itemize}

  \item  $T  \subset  \obj{E}$  be  a set  of  edges  defining  people's trajectories.

  \item $P$ be a  set of patterns, each defining an  area where people behaving
  in a specific way  are likely to be found, plus  an empty pattern $\emptyset$
  used to describe unusual behaviors. Formally speaking, patterns are functions
  that associate to a trajectory with an arbitrary number of edges a score that
  denotes how likely it is to correspond  to that specific pattern, as shown in
  Section~\ref{sec:objFunc}.

  \item $A$ be a set of assignments  of individual detections in $\obj{D}$ into
    patterns, that  is, a mapping $A:  D \rightarrow \{1, \ldots,  N_p\}$, where
    $N_p$ is the total number of patterns.

\end{itemize}
Each trajectory $t \in T$ must go through detections via allowable transitions,
begin at $I$, and end at $O$. Here we abuse the notation $t \in T$ to show that
all  edges  $(I,  t_1),  (t_1,  t_2), \cdots,  (t_{|t|},  O)$  from  trajectory
$t=(t_1,\cdots,t_{|t|})$  belong to  $T$. Furthermore,  since we  only consider
high-confidence  detections,  each  one  must belong  to  \textbf{exactly}  one
trajectory.  In practice,  this means  that  potential false  positives end  up
being  assigned to  the empty  behavior  $\emptyset$ and  can be  removed as  a
post-processing  step. Whether  to  do this  or  not is  governed  by a  binary
indicator $R_e$ selected during training process.  In other words, the edges in
$T$ must  be such that  for each detection there  is exactly one  selected edge
coming in and one going out, which we can write as
\begin{equation}
  \label{eq:ENConstr1}
\forall j \in  \obj{D} , \exists! i \in  \obj{V}, k \in \obj{V}: (i,  j) \in T
\land (j, k) \in  T\; .
\end{equation}
Since  all detections  that are  grouped into  the same  trajectory $T$  must be
assigned to the same pattern, we must have
\begin{equation}
  \label{eq:ENConstr2}
\forall (i, j) \in T:  (i \in \obj{D} \land j \in \obj{D}) \Rightarrow A(i) = A(j)\; .
\end{equation}
In  our  implementation, each  pattern  $p  \in  P  \backslash \emptyset  $  is
defined  by  a  trajectory  that  serves  as  a  centerline  and  a  width,  as
depicted  by Fig.~\ref{fig:graph}(c)  and~\ref{fig:n-and-m-functions}. However,
the optimization  schemes we  will describe  in Sections~\ref{sec:trajectories}
and~\ref{sec:patterns} do  not depend on  this specific representation  and any
other convenient one could have been used instead.

\subsection{Building the Graph}
\label{sec:graphBuilding}

To  build the  graph we  use  as input  the  output of  another algorithm  that
produces trajectories  that we want to  improve. We take the  set of detections
along these  trajectories to  be our  high-confidence detections  $\obj{D}$ and
therefore the nodes  of our graph. We  take the edges $\obj{E}$ to  be pairs of
nodes that are  either {\it i)} consecutive in the  original trajectories, {\it
ii)} within  ground plane distance  $D_1$ of  each other in  successive frames,
{\it iii)}  the endings  and beginnings of  input trajectories  within distance
$D_2$ and within $D_t$  frames, {\it iv)} or whose first node  is $I$ or second
node  is  $O$. In  other  words,  to allow  the  creation  of new  trajectories
and  recover  from identity  switches,  fragmentation,  and incorrectly  merged
trajectories, we  introduce edges not  only for consecutive points  in existing
ones but also to connect neighboring trajectories.

\comment{\PF{Add back the relevant sentences  about additional knowledge and edges. There
  should be somewhere a list or table of all parameters (e.g. $D_1$) set by hand
  and all those that are learned}.}

\comment{  are  constructed  using  the  following
intuitive rules:  {\it i)} Each pair  of detections in the  neigibouring frames
within the are connected by an edge to fix the possible swaps
between  identities; {\it  ii)} If  we are  given information  about the  input
trajectories,  pairs  of  last  /  first detections  in  the  trajectories  are
connected if  they are within  the distance of $D_2$  and at most  $D_t$ frames
apart  to fix  possible fragmented  trajectories; {\it  iii)} If  we are  given
information about input trajectories, transitions in them are added as edges of
detection graph to respect the results  of the input tracking method; {\it iv)}
Edges from $I$ to all detections and  from all detections to $O$ are present in
the detection graph to fix possible merged trajectories.}

\subsection{Objective Function}
\label{sec:objFunc}

Our  goal is to find the most  likely trajectories formed by transitions in $T^*$, patterns
$P^*$, and mapping linking one to the  other $A^*$ given the image information and
any {\it  a priori}  knowledge we  have. In  particular, given a set of patterns
$P^*$, we will look for the best set of trajectories that match these
patterns. Conversely, given a set of  known trajectories $T^*$, we will learn a
set of patterns, as discussed in Section~\ref{sec:computing}.

To formulate these  searches in terms of an optimization  problem, we introduce
an objective function $C(T,P,A)$ that reflects  how likely it is to observe the
objects moving along the trajectories defined by $T$, each one corresponding to
a  pattern from  $P=\{p_1\cdots,p_{N_p}\}$  given the  assignment  $A$. Ideally,  $C$  should be  the
proportion of trajectories that correctly follow the assigned patterns. To compute it in
practice, we take our inspiration from the \MOTA{} and \IDF{} scores described
in  Section~\ref{sec:metrics}. They  are  written  in terms  of  ratios of  the
lengths  of  trajectory  fragments  that  follow  the  ground  truth  to  total
trajectory lengths. We therefore take our objective function to be

\comment{To measure  how well a trajectory
follows  a pattern,  we follow  the approach  similar to  the one  used in  MOT
metrics  , which  measure how  well the  trajectory fits  to
ground truth. In  particular, for each edge  of the trajectory we  want to know
whether it matches the selected pattern or  not, and express the final score as
a ratio  of matched  edges to the  total number of  edges used.  However, since
people can move with  varying speed or stop for prolonged  momemnts of time, we
are actually more  interested in the ratio of the  total {\it length} travelled
along the pattern. Therefore, we use the following cost function:}
\vspace{-0.4cm}
\begin{small}
\begin{eqnarray}
  C(T,P,A) \mbox{\hspace{-0.1cm}} & = \mbox{\hspace{-0.1cm}} &
  {\sum\limits_{t \in T} M(t,p_{A(t_1)})
  \over
  \sum\limits_{t \in T} N(t,p_{A(t_1)})}, \label{eq:costFun} \\
  N(t, p) \mbox{\hspace{-0.1cm}} & = \mbox{\hspace{-0.1cm}}& n(I, t_1, p) + n(t_{|t|}, O, p) + \mbox{\hspace{-0.3cm}} \sum\limits_{
    1 \leq j \leq |t| - 1} \mbox{\hspace{-0.3cm}} n(t_j, t_{j + 1}, p), \nonumber \\
  M(t, p) \mbox{\hspace{-0.1cm}} & = \mbox{\hspace{-0.1cm}}& m(I, t_1, p) + m(t_{|t|}, O, p) + \mbox{\hspace{-0.5cm}} \sum\limits_{
    1 \leq j \leq |t| - 1} \mbox{\hspace{-0.3cm}}m(t_j, t_{j + 1}, p), \nonumber
\end{eqnarray}
\end{small}
where $n(i, j, p)$ is  the sum of the total length  of edge $(i,
j)$ and  of the  length of  the corresponding  pattern centerline,  while $m(i,
j,  p)$  is the sum of lengths of aligned parts of the  pattern  and  the  edge.
Fig.~\ref{fig:n-and-m-functions} illustrates  this computation and we  give the
mathematical definitions  of $m$ and  $n$ in  the supplementary material.  As a
result, $N(t, p)$ is the sum of  the lengths of trajectory and assigned pattern
while $M(t, p)$ measures the length of parts of trajectory and pattern that are
aligned with  each other. Note  that the definition of  Eq.~\eqref{eq:costFun} is
very close to  that of the metric \IDF{}  introduced in Sec.~\ref{sec:metrics}.
It  is largest  when  each person  follows  a  single pattern  for  as long  as
possible. This  penalizes identity switches  because the trajectories  that are
erroneously merged, fragmented,  or jump between people are  unlikely to follow
any of such pattern.

\input{figures/n-and-m-functions}

\comment{ \pf{where $m(i, j,  p)$ is the length of the path  from that is within
    the pattern plus the length of  the pattern centerline, $n(i, j, p)$ denotes
    the total  length of  the edge $(i,  j)$ plus the  length of  an appropriate
    transition in the pattern. }

Conceptually, we define  functions $m$ and $n$ so that,  for any trajectory $t$
and  pattern  $p$, the  value  of  $n(t,  p) =  n(I,  t_1)  + n(t_{|t|},  O)  +
\sum\limits_{j \in  \{1, \dots, |t| -  1\}}n(t_j, t_{j + 1})$  approximates the
sum of the total lengths of both  pattern and trajectory, while $m(t, p) = m(I,
t_1) + m(t_{|t|}, O)  + \sum\limits_{j \in \{1, \dots, |t|  - 1\}}m(t_j, t_{j +
1})$  approximates  the  overlap  length between  pattern  and  trajectory,  as
illustrated by Fig.~\ref{fig:n-and-m-functions}.}

\comment{\am{The  key
properties of the defined function are: {\it i)} It scores the whole trajectory
and even set of trajectories,  rather than each independent transition, freeing
us  from a  Markovian  assumption  of people  movement;  {\it  ii)} It  closely
resembles the metrics  such as \MOTA{} and \IDF{} used  to evaluate the quality
of people tracking, which are also fractions with the number of matched objects
in numerator and total number of objects in denominator; {\it iii)} It accounts
for  trajectories of  people of  varying  speed {\it  iv)}  As we  will see  in
Sec.~\ref{sec:methodTrajectories}, it can be  optimized efficiently.} We define
$m$ and $n$ more precisely in the supplementary material.}

In Eq.~\eqref{eq:costFun},  we did not  explicitly account  for the fact  that the
first vertex $i$ of some edges can  be the special entrance vertex, which is not
assigned to any  behavior. When this happens we simply  use the pattern assigned
to the second  vertex $j$.  From now  on, we will replace $A(i)$  by $A(i,j)$ to
denote this behavior.  We also adapt the definitions of  $m$ and $n$ accordingly
to properly handle those special edges.


%% file: figures/n-and-m-functions.tex

\begin{figure}[t]
\begin{center}
\hspace{-1pt}
\includegraphics[trim={4cm 11.8cm 16cm 6cm},clip,width=0.5\textwidth]{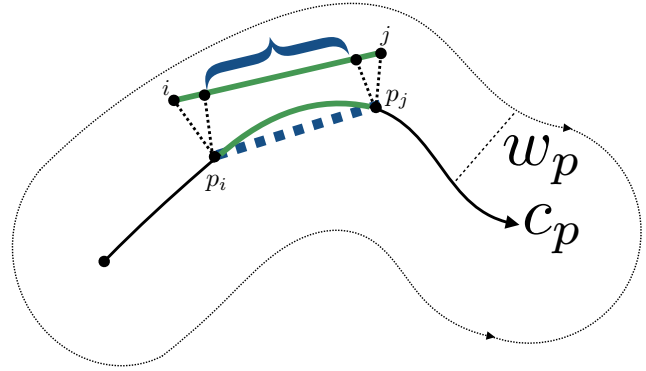}
\end{center}
\vspace{-0.5cm}

\caption{For a  pattern $p$ defined by  centerline $c_p$, shown as a thick black
line, with width $w_p$, and an edge $(i, j)$, we compute functions $n(i, j, p)$
and $m(i, j, p)$ introduced in Section~\ref{sec:objFunc} and shown in green and
blue, respectively, as  follows: $n(i, j, p)$  is the total length  of the edge
and the  corresponding length of  the pattern centerline, measured  between the
points $p_i$ and $p_j$,  which are the points on the  centerline closest to $i$
and  $j$. If  both $i$  and $j$  are within  the pattern  width $w_p$  from the
centerline, we take $m(i, j, p)$ to be  the sum of two terms: the length in the
pattern along the edge, that is, the  distance between $p_i$ and $p_j$, plus the
length in the edge along the pattern,  that is, the length of the projection of
$(p_i, p_j)$ onto the line connecting $i$ and $j$. Otherwise $m(i, j,p) = 0$ to
penalize the deviation from the pattern.}
\label{fig:n-and-m-functions}
\vspace{-0.5cm}
\end{figure}

\comment{
\begin{figure}[t]
\begin{center}
\hspace{-1pt}
\includegraphics[trim={11cm 8.9cm 10cm 8.9cm},clip,width=0.5\textwidth]{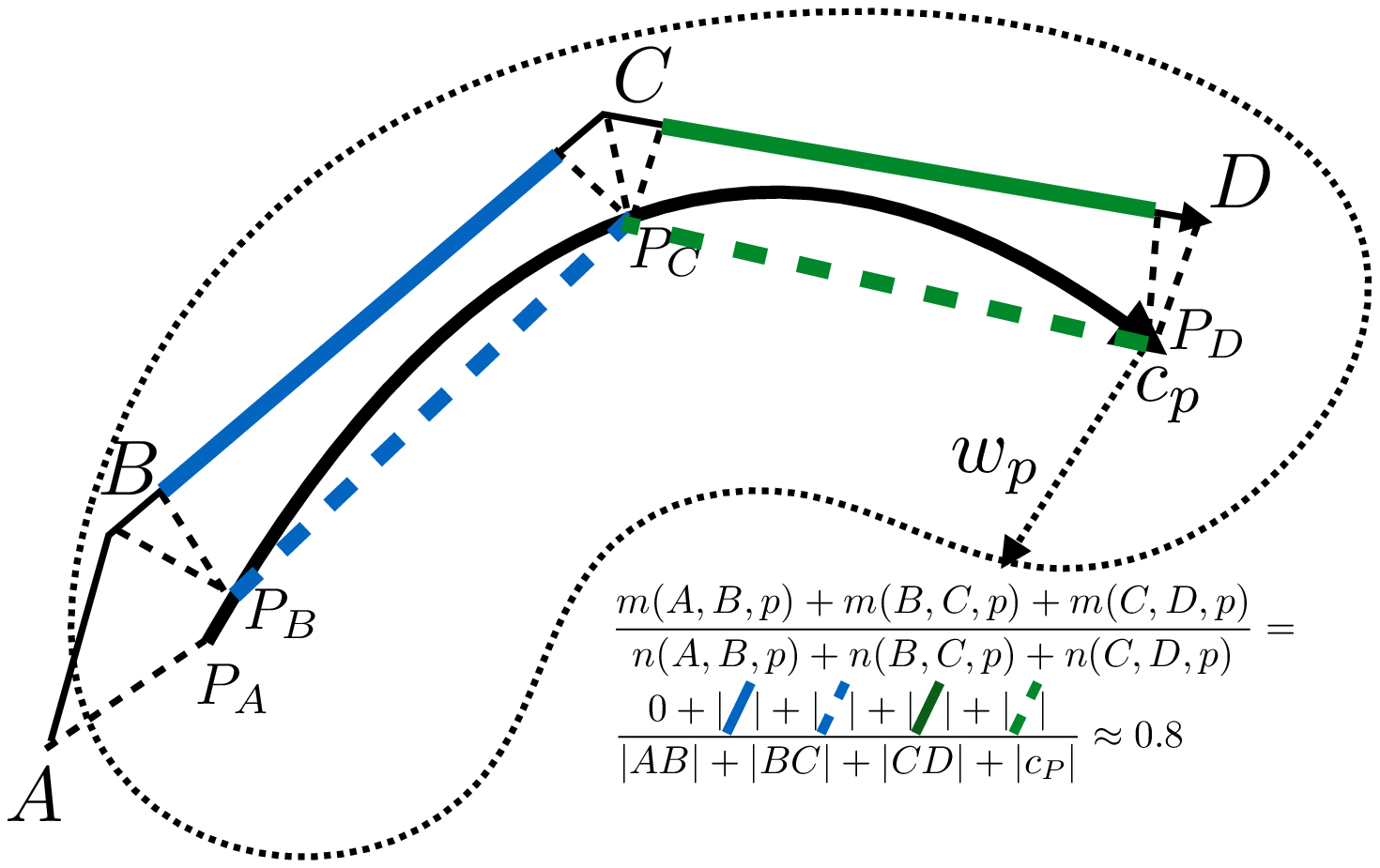}
\end{center}

\caption{For a  pattern $p$ defined by  centerline $c_p$, shown as  thick black
line, and width  $w_p$ and a trajectory  defined by edges $(A, B),  (B, C), (C,
D)$, the cost  function is expressed as a  ratio of sum of $m$ over  sum of $n$
for all the edges. For example, for  edge $(B, C)$, $m(B, C, p)$ corresponds to
the distance in pattern travelled along $(B, C)$, marked with blue dashed line,
plus the  distance in $(B, C)$  travelled along the pattern,  marked with solid
blue line. To compute the former, we find the distance between the points $P_B$
and $P_C$ on the  pattern closest to $B$ and $C$, and to  compute the former we
project them back  onto the line connecting $B$ and  $C$. Similar situation for
the  edge $(C,  D)$  is shown  in  green. For  the  edge $(A,  B)$,  it is  not
completely within  the distance $w_p$ from  the pattern, which is  why $m(A, B,
p)=0$. $n(B, C, p)$  is computed as a sum of the edge  length $(B, C)$ plus the
length of the pattern between $P_B$ and $P_C$, which is why the sum of function
$n$ for all edges  yields the total length of trajectory  plus the total length
of the pattern.}
\label{fig:n-and-m-functions} \end{figure}
}

%% file: tracking.tex

\section{Computing Trajectories and Patterns}
\label{sec:computing}

In  this  section,  we  describe  how  we use  the  objective  function  $C$  of
Eq.~\eqref{eq:costFun} to  compute trajectories given patterns  and patterns given
trajectories.   The resulting  procedures will  be  the building  blocks of  our
complete MOT algorithm, as described in Section~\ref{sec:mot}.

\subsection{Trajectories}
\label{sec:trajectories}

Let us assume that we are given a precomputed set of patterns $P^*$, then we look for
trajectories and corresponding assignment as
\vspace{-0.3cm}
\begin{equation}
  T^*, A^* = \argmax_{T,A} C(T,P^*,A) \; .
\end{equation}
To solve  this problem,  we treat  the motion of  people through  the detection
graph $\obj{G}$ introduced in Section~\ref{sec:graph}  as a flow. Let $o^p_{ij}
\in \{0,1\}$ be  the number of people  transitioning from node $i$ to  $j$ in a
trajectory $T$ assigned to pattern $p \in P^*$. It relates to $P^*$ and
$T$ as follows:
\begin{equation}
  o^p_{ij} = \mathbb{I}(((i, j) \in T) \land (P^*_{A(i,j)} = p)) \; .
\end{equation}
Using these new binary  variables, we reformulate constraints \eqref{eq:ENConstr1}
and \eqref{eq:ENConstr2} as
%
\vspace{-0.1cm}
\begin{eqnarray}
  \forall   i  \in   \obj{D}\cup{O}  \sum\limits_{(i,j)   \in  \obj{E},   p  \in
    P^*}o^p_{ij} & = & 1 \;, \label{eq:flowConstraints} \\
  \forall j \in \obj{D}, p \in  P^* \sum\limits_{(i,j) \in \obj{E}}o^p_{ij} &
  = & \sum\limits_{(j,k) \in \obj{E}}o^p_{jk} \;. \nonumber
\end{eqnarray}
This lets us rewrite our cost function as
\begin{equation}
  \label{eq:flowCostFun}
  C(T,P^*,A) = {\sum\limits_{(i, j) \in T, p \in P^*} m(i,j,p) o^p_{ij}\over\sum\limits_{(i, j) \in T, p \in P^*} n(i,j,p) o^p_{ij}} \; ,
\end{equation}
which we maximize  with respect to the flow variables  $o^p_{ij}$ subject to the
two constraints of \eqref{eq:flowConstraints}.   This is an integer-fractional
program,  which could  be  transformed into  a Linear  Program~\cite{Charnes62}.
However,  solving it  would produce  non-integer values  that would  need to  be
rounded. To avoid  this we propose a scheme based  on the following observation:
Maximizing $a(x) \over b(x)$ with respect  to $x$ when $b(x)$ is always positive
can be  achieved by  finding the  largest $\alpha$ such  that an  $x$ satisfying
$a(x) - \alpha b(x) \ge 0$ can  be found.  Furthermore, $\alpha$ can be found by
binary   search.    We   therefore   take   $a$   to   be   the   numerator   or
Eq.~\eqref{eq:flowCostFun}, $b$ its denominator, and  $x$ the vector of $o^p_{ij}$
variables. In practice,  given a  specific value  of $\alpha$,  we do  this by
  running  a  Integer  Linear  Program solver~\cite{Gurobi}  until  it  finds  a
  feasible  solution. When  $\alpha$ reaches  its maximum  possible value,  that
  feasible solution  is also the  optimal one. We  provide more details  in the
supplementary material and a version of our our code is publicly available
\footnote{https://github.com/maksay/ptrack\_cpp}.

%% file: patterns.tex

\subsection{Patterns}
\label{sec:patterns}

In the previous section, we assumed the  patterns known and used them to compute
trajectories. Here, we reverse  the roles.  Let us assume we are  given a set of
trajectories $T^*$. We learn the patterns and corresponding assignments as
\vspace{-0.1cm}
\begin{eqnarray}
  P^*, A^* & = & \argmax_{P,A} C(T^*,P,A) \; , \label{eq:patternEq} \\
  \mbox{subject to} && P \subset \obj{P}, |P| \le \alpha_p, \sum\limits_{p \in P}M(p) \le \alpha_c \nonumber \; ,
\end{eqnarray}
where  $\alpha_c,  \alpha_p$  are  thresholds and  $M:P\to  \mathbb{R}^+$.  The
purpose  of  the  additional  constraints  is  to  limit  both  the  number  of
patterns being  used and their spatial  extent to prevent over-fitting.  In our
implementation, we  take $M(p)  = l_p w_p$,  where $l_p$ is  the length  of the
pattern  centerline  and  $w_p$  is  its  width. $\obj{P}$  is  a  set  of  all
admissible patterns, which we construct  by combining all possible ground-truth
trajectories as centerlines  with each width from a predefined  set of possible
pattern widths.
\comment{
\pf{This  maximization  problem  can  be   solved  simply  by  reformulating  it
  as}\PF{What's wrong with my formulation?}
\comment{Instead of solving the maximization problem above, we will minimize $M$:}
\begin{equation}
  \label{eq:patternConstr}
  (P^*, A^*) =  \argmin\limits_{P, A: C(T^*, P, A)  \ge \alpha_C} \sum\limits_{p
    \in P}M(p) \; .
\end{equation}
This amounts  to looking for the  smallest set of  paterns in terms of  $M$ that
explains  at  least $\alpha_C$  percent  of  trajectories  that  we see  in  the
ground-truth   data    $T^*$.   \PF{Really?}}

To solve  the problem  of Eq.~\eqref{eq:patternEq}, we  look for  an assignment
between our  known ground  truth trajectories $T^*$  and all  possible patterns
$\obj{P}$ and  retain only patterns associated  to at least one  trajectory. To
this end, we  introduce auxiliary variables $a_{tp}$  describing the assignment
$A^*: T^*  \rightarrow \obj{P}$, and variables  $b_p$ denoting if at  least one
trajectory is matched to pattern $p$. Formally, this can be written as
\vspace{-0.3cm}
\begin{eqnarray}
  a_{tp}&\in&    \{0,1\}
  \; , \forall t \in T^*, p \in \obj{P} \;, \nonumber\\
  b_{p}&\in&\{0,1\}\; ,  \forall p \in \obj{P} \; , \nonumber\\
  \sum\limits_{p    \in   \obj{P}}a_{tp}&   =&
  1 \;, \forall t \in T^* \; , \label{eq:matchConstraints}\\
  a_{tp}&  \le &  b_p \; , \forall t \in T^*, p \in
  \obj{P} \; .\nonumber
\end{eqnarray}
Given  that  $C$  is  defined  as  the  fraction  from  Eq.~\eqref{eq:costFun}, we
  use the optimization scheme similar to one described in Sec.~\ref{sec:trajectories},
  where we do binary search to find the optimal value of $\alpha$ such that there exists a
feasible solution for constraints of~\eqref{eq:matchConstraints} and the following:
\vspace{-0.1cm}
\begin{align}
\label{eq:matchFractionConstr}
& \sum\limits_{t \in T^*}\sum\limits_{p \in \obj{P}} (m(t, p) - \alpha n(t, p)) a_{tp} \ge 0 \;, \nonumber \\
& \sum\limits_{p \in \obj{P}}b_p \le \alpha_p \;,  \\
& \sum\limits_{p \in \obj{P}}b_p M(p) \le \alpha_c \;. \nonumber
\end{align}
\comment{and  the  minimization  problem  of Eq.~\eqref{eq:patternEq}  reduces  to  the
\pf{Integer Linear Program}
\begin{equation}
\label{eq:finalEq}
\text{minimize}\displaystyle\sum\limits_{p \in \obj{P}}b_p M(p)
\end{equation}
under the constraints of Eqs.~\eqref{eq:matchConstraints} and~\eqref{eq:matchFractionConstr}.}

%% file: mot.tex

\section{Non-Markovian Multiple Object Tracking}
\label{sec:mot}

Given that we  can learn patterns from  a set trajectories, we  can now enforce
long-range  behavioral  patterns  when  linking  a set  of  detections  in  two
different manners. \am{This is in  contrast to traditional approaches enforcing
local smoothness constraints, Markovian in their essence.}

If  annotated ground-truth  trajectories $T^*$  are available,  we use  them to
learn the patterns as described in Sec.~\ref{sec:patterns}. Then, at test
time, we use the linking procedure of Sec.~\ref{sec:trajectories}.

If no such training data is available,  we can run an E-M-style procedure, very
similar to the Baum-Welch algorithm~\cite{Jelinek75}  for HMMs: we start from a
set of  trajectories computed using a  standard algorithm, and from  there, use
trajectories to  compute a  set of patterns,  then use the  set of  patterns to
compute trajectories,  and iterate. We  will see  that, in practice,  this yields
results that are almost indistinguishable in  terms of accuracy but much slower
because we have to run through many iterations.

More specifically, each iteration of our unsupervised approach involves {\it i)
} finding a set of patterns $P^{i}$ given a set of trajectories $T^{i - 1}$,
{\it  ii) }finding  a set  of trajectories  $T^{i}$ given  a set  of patterns
$P^{i}$,        as       described        in       Sec.~\ref{sec:patterns}
and~\ref{sec:trajectories}.

In  practice,  for a  given  $\alpha_c$,  this  scheme  converges after  a  few
iterations. Since $\alpha_c$  is unknown {\it a priori}, we  start with a small
$\alpha_c$,  perform  5  iterations,  increase  $\alpha_c$,  and  repeat  until
reaching  a predefined  number of  patterns.  To select  the best  trajectories
without reference to ground truth, we define
\begin{equation*}
  \begin{small}
\widetilde{\text{\IDF}}(T^i) = {1\over 2}(C(T_1^i, P_2^i,  A_{T_1^i \to P_2^i}) + C(T_2^i, P_1^i,
A_{T_2^i \to P_1^i})) \; ,
  \end{small}
\label{eq:idfTilde}
\end{equation*}
where  $T_1^i$ and  $T_2^i$ are  time-disjoint  subsets of  $T^i$, $P_1^i$  and
$P_2^i$ are  patterns learned from  $T_1^i$ and $T_2^i$. $A_{T_1^i  \to P_2^i}$
and $A_{T_2^i \to P_1^i}$ are such  assignments of trajectories to the patterns
learned on another subset that maximize $\widetilde{\text{\IDF}}(T^i)$.

In  effect,  $\widetilde{\text{\IDF}}$  is  a  valid  proxy  for  \IDF{}  due
to  the  many  similarities  between  our cost  function  and  \IDF{}  outlined
in  Sec.~\ref{sec:objFunc}.  In  the  end,  we  select  the  trajectories  that
maximize  $\widetilde{\text{\IDF}}$. Using  such cross-validation  to pick  the
best solution in E-M models is justified in~\cite{EST10}.

%% file: results.tex
\section{Results}
\label{sec:results}

\comment{We either  use the  trajectories  to learn  patterns or  produce
improved   trajectories,   as  described   in   Sections~\ref{sec:trajectories}
and~\ref{sec:patterns}.}

In this  section, we demonstrate the  effectiveness of our approach  on several
datasets, using both simple and sophisticated approaches to produce the initial
trajectories.
(Recall  from  Section~\ref{sec:graphBuilding}  that  we  build  our  detection
graphs from  the output  of another tracking  algorithm.)
In the remainder of this  section, we first describe the datasets
and the  tracking algorithms we  rely on to build  the initial graphs.  We then
discuss  the evaluation  metrics  and the  experimental  protocol. Finally,  we
present our experimental results.

\subsection{Datasets}
\label{sec:datasets}

\input{figures/datasets}

\noindent We use the four datasets listed in Tab.~\ref{tab:datasets}. They are:

\Towncentre{}.   A sequence  from  the 2DMOT2015  benchmark  featuring a  lively
Zurich street where people walk in different directions.

\ETH{}   and   \Hotel.   Sequences    from   the   BIWI   Walking   Pedestrians
dataset~\cite{Pellegrini09} that were originally used to model social behavior.
In  these datasets,  using image  and  appearance information  for tracking  is
difficult,  due to  recordings with  a top  view camera  and low  visibility in
\ETH{} dataset.

\Station{}. A one hour-long recording of  Grand Central station in New York with
several  thousands   of  annotated  pedestrian  tracks~\cite{Zhou12a}.   It  was
originally  used  for  trajectory  prediction  for  moving  crowds.  \comment{As
  in~\cite{Zhou12a}, we detect  the patterns only for moving  subjects, which is
  why we preprocessed  the dataset to treat the trajectory  of people before and
  after    long   stationary    pauses   (\eg    queues)   as    two   different
  trajectories.\AM{Since we anyway use for  evaluation periods of 1 minute long,
    we break each long trajectory in every dataset into several independent ones
    for  the  evaluation  purposes.   So  maybe we  don't  need  to  mention  it
    here.}\PF{If you don't need to mention it, don't}}

  \comment{\item \PTZ{}. Outdoor sequences of a crowded university campus seen from three
    different  viewpoints~\cite{Possegger12a}.  We  treated each  one separately
    and annotated the  dataset ourselves.  \am{We provide  averaged results here
  and individual breakdown in supplementary materials.}}

These four datasets share the following characteristics: {\it i)} They feature
real-life behaviors as opposed to random  and unrealistic motions acquired in a
lab setting; {\it ii)} The frame rate is at most 5 frames per second, which is
realisitic for outdoor  surveillance setups but makes  tracking more difficult;
{\it iii)} They are all single-camera but  the shape of the ground surface can
be estimated from the bottom of the  bounding boxes, which makes it possible to
reason in a simulated  top view as we do. In other words,  they are well suited
to test our approach in challenging conditions.

\subsection{Baselines}
\label{sec:baselines}

As  discussed  in  Section~\ref{sec:graphBuilding},  we use  as  input  to  our
system  trajectories  produced   by  recent  MOT  algorithms,   some  of  which
exploit  image and appearance  information   and  some  of   which  do   not.  In
Section~\ref{subsec:experiments},  we  will  show  that  imposing  our  pattern
constraints  systematically results  in  an improvement  over these  baselines,
which we list below.
\vspace{-4mm}
\paragraph{\MDP~\cite{Xiang15}} formulates  MOT as  decision making  in a Markov
Decision  Process (MDP)  framework.  Learning to  associate  data correctly  is
equivalent  to  learning  an  MDP  policy and  is  done  through  reinforcement
learning.  At  the time  of  writing,  this  was the  highest-ranking  approach
(in  terms   of  \MOTA{})  with   publicly  available  implementation   on  the
2DMOT2015~\cite{Leal-Taixe15} benchmark.
\vspace{-4mm}
\paragraph{\SORT~\cite{Bewley16}}  is  a   real-time  Kalman  filter-based  MOT
approach. At the time of writing, this was the second highest-ranking approach
on 2DMOT2015 benchmark.
\vspace{-4mm}
\paragraph{\RNN~\cite{Milan16}} is  a recent attempt at  using recurrent neural
networks to predict the motion of multiple people and perform MOT in real time.
In does  not require any  appearance information,  only the coordinates  of the
bounding boxes.  In the  presented results, this  approach outperforms  all the
other methods that do not use image and appearance information.
\vspace{-8mm}
\paragraph{\KSP~\cite{Berclaz11}} is a simple approach  to MOT that formulates
the MOT  problem as finding  K Shortest Paths  in the detection  graph, without
using image or appearance information.
\vspace{-4mm}
\paragraph{\bf  2DMOT2015   Top  Scoring  Methods}~\cite{Choi15,Sanchez-Matilla16,Wang16b,Keuper16,Yang16,Kim15,Wang16c,Hong16}
 to which  we will refer by  the name that appears  in the official
scoreboard~\cite{Leal-Taixe15}. This will allow us to show that our approach is
widely applicable.

Top  scoring MOT  methods from  the  2DMOT2015 benchmark  on the  \Towncentre{}
dataset rely  on a people  detector that is  not always publicly  available. We
therefore used  their output  to build  the detection  graph, and  report their
results  only on  \Towncentre{} dataset.  For  all others,  the available  code
accepts a  set of  initial detections  as input. To  compute them,  we obtained
background subtraction  by subtracting the  median image. We used  the publicly
available POM  algorithm of~\cite{Fleuret08a} on  the resulting binary  maps to
produce probabilities of presence in various ground locations and we kept those
for which  the probability was greater  than 0.5. This proved  effective on all
our datasets. For comparison purposes, we  also tried using SVMs trained on HOG
features~\cite{Dalal05} and deformable part models~\cite{Felzenszwalb10}. While
their performance  was roughly similarly  to that  of POM on  \Towncentre{}, it
proved much worse when the people are far away or seen from above.

\subsection{Experimental Protocol}
\label{sec:protocol}

On  the  \Station{}  dataset,  which  is   long  and  features  more  than  100
people  per  minute,  we  tested   on  1-minute  subsequences  and  trained  on a
non-overlapping 5-minute subsequence. \am{We also limited the optimization time
for  solving Eq.~\ref{eq:flowCostFun}  to 10  minutes per  iteration of  binary
search.} On  all other  datasets, we tested  on 1-minute  subsequences, trained
on  the  remainder  and  did  not  limit  optimization  time.  To  prevent  any
interaction  between  the  training  and  testing data,  we  removed  from  the
ground  truth  training  data  all  incomplete  trajectories  to  guarantee  no
overlap  with  the  testing  data.  The remaining  trajectories  were  used  to
learn  the patterns  of  Section~\ref{sec:patterns} and  choose  the values  of
the  parameters  $D_1$,  $D_2$,  $D_t$ controling  the  construction  of  edges
in  the  tracking  graph,  $R_e$ controling  whether  to  discard  trajectories
assigned  to  no  pattern,   and  $\alpha_c$,  $\alpha_p$  regularizing  number
and  width  of  patterns,  introduced  in  Sections~\ref{sec:trajectories}  and
~\ref{sec:patterns}.  It is  done by  performing  a grid  search and  selecting
values that yield  the best possible \IDF{} score in  cross-validation. To keep
the search tractable,  we always started from a default  set of values $D_1=2m,
D_2=4m, D_t=2s,  \mathbb{I}(R_e) = 1, \alpha_p=5,  \alpha_c=0.3 \alpha_p \times
\text{area  for tracking}$,  and explored  neighboring values  in the  6D grid.
We  do  the same  exploration  when  running  iterative  scheme to  select  the
optimal value  of $\widetilde{\text{\IDF}}$ in unsupervised  setup described in
Section~\ref{sec:mot}.

For   the  sake   of  fairness,   we   trained  the   trainable  baselines   of
Section~\ref{sec:baselines}, that is \MDP{} and \RNN{}, similarly and using the
same data.  However, for \RNN{} we  obtained better results using  the provided
model, pre-trained on the 2DMOT2015 training data, and we report these results.

We combined the results from all test segments to obtain the overall metrics on
each dataset.  Since for  some approaches we  only had results  in the  form of
bounding boxes  and had to  estimate the ground  plane location based  on that,
this often  resulted in  large errors  further away from  the camera.  For this
reason, we evaluated \MOTA{} and \IDF{}  assuming that a match happens when the
reported location  is at most  at 3 meters from  the ground truth  location. We
also provide results  for the traditional distance of 1  meter in supplementary
material and they  are similar in terms of method  ordering. For the \Station{}
dataset, we did not have  the information about the true size  of the floor area
(we estimated the  homography between the image and ground  plane) which is why
we used as distance 0.1 of the size of the tracking area.

\input{figures/towncentre}

\input{figures/barplot}

\subsection{Experiments}
\label{subsec:experiments}

\input{figures/pictures}
\input{figures/iterative}

 We first  show that our approach  consistently improves the output  of all the
 baselines using initial people detections obtained  as described at the end of
 Section~\ref{sec:baselines}. Then,  to gauge  what our approach  could achieve
 given perfect  detections, we perform  this comparison again but  using ground
 truth detections instead. Finally, we  discuss the computational complexity of
 our approach.

\vspace{-0.3cm}
\paragraph{Improving Baseline Results.}

In terms of the \IDF{} metric,  as can be seen is Fig.~\ref{fig:towncentre}, our
Supervised method improves most of the  tracking results except one that remains
Unchanged on \Towncentre{}. The same can  be said of the unsupervised version of
Our   method  except   for   one  result   that  it   degrades   by  0.01.    In
Tab.~\ref{fig:barplot},  we average  these  results over  all  datasets and  can
Observe  a marked  average  improvement for  all  methods we  could  run on  all
Datasets both  in the supervised and  unsupervised cases. As could  be expected,
The improvements  in terms of \MOTA{}  are less clear since  our method modifies
The set of input detections  minimally.  Fig.~\ref{fig:pictures} depicts some of
The results and we provide detailed breakdowns in the supplementary material.

\paragraph{Evaluation  on  Ground Truth  Detections.}  For  all baselines  that
Accept a list of  detections as input, and for which the  code is available, we
Reran the  same experiment using the  ground truth detections instead  of those
Computed by  the POM algorithm~\cite{Fleuret08a}  as before.  This is a  way to
Evaluate the performance of the linking  procedure independently of that of the
Detections. It reflects the theoretical maximum  that can be reached by all the
Approaches  we compare,  including our  own. From  Tables~\ref{tab:eval-gt-idf}
And~\ref{tab:eval-gt-mota}, we observe that our  approach performs very well in
Such setting.

\input{figures/eval-gt-idf}
\input{figures/eval-gt-mota}
\input{figures/time.tex}

\paragraph{Computation time}

The number  of variables in  our optimization  problem grows linearly  with the
Length of the  batch and number of patterns, and  superlinearly with the number
Of people per frame (as the  number of possible connections between people). As
Shown by Tab.~\ref{tab:time}, for not too crowded datasets without large number
Of patterns  our approach  is able to  process a minute  of input  frames under
A  minute.  Pattern fitting  scales  quadratically  with  the number  of  given
Ground-truth trajectories  and runs in  less than  10 minutes for  all datasets
Except \Station{}. More details can be found in supplementary materials.

%% file: figures/datasets.tex
\begin{table}[!b]
\vspace{-0.25cm}
\begin{tabular}{|c|c|c|c|}
\hline
Name          & Annotated length, s & FPS  & Trajectories \\
\hline
\Towncentre{} & 180                 & 2.5  & 246 \\
\hline
\ETH{}        & 360                 & 4.16 & 352 \\
\comment{\hline
\PTZ{}        & 360$\times$3        & 1    & 130 + 69 + 142 \\}
\hline
\Hotel{}      & 390                 & 2.5  & 175 \\
\hline
\Station{}    & 3900                & 1.25 & 12362 \\
\hline
\end{tabular}
\caption{Dataset  statistics. The  number of  trajectories is  calculated as  a
total sum of number of trajectories in each test set on which we evaluated. All
test sets were approximately 1min long.}
\label{tab:datasets}
\end{table}

%% file: figures/towncentre.tex
\begin{figure}[!t]
\begin{center}
\hspace{-1pt}
\begin{tabular}{c}
  \hspace{-0.4cm}\includegraphics[trim={3cm 0.0cm 6.0cm 0cm},clip,width=0.49\textwidth]{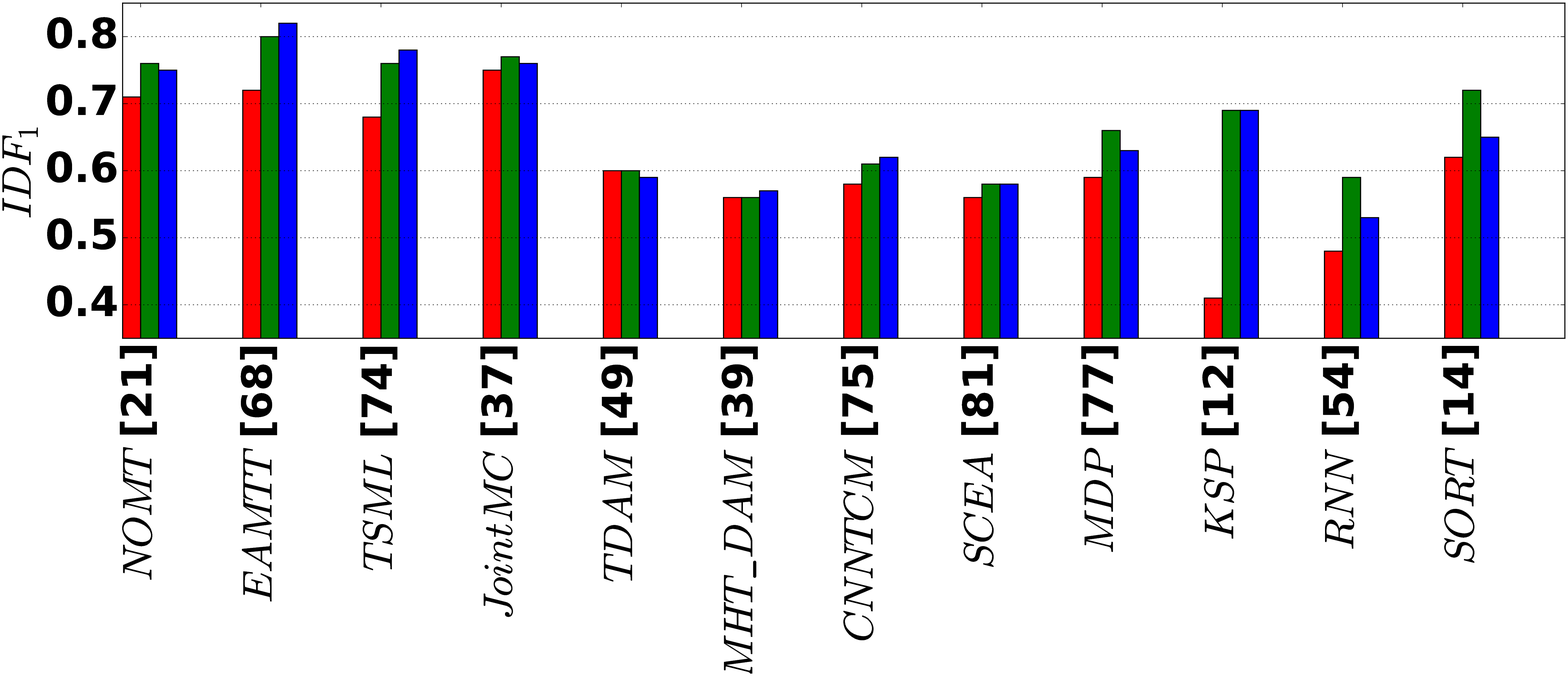} \\
  \hspace{-0.4cm}\includegraphics[trim={3cm 0cm 9.0cm 1.6cm},clip,width=0.49\textwidth]{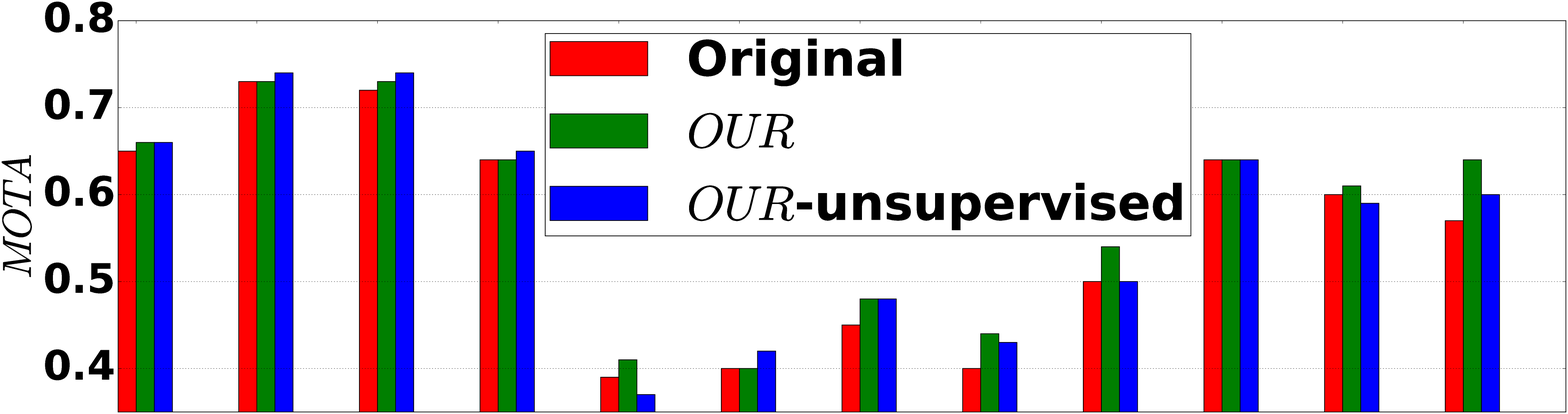}
\end{tabular}
\end{center}
\vspace{-0.6cm}
\caption{\IDF{}  and \MOTA{}  scores for  various methods  on the  \Towncentre{}
  dataset. Our  approach almost  always improves \IDF{}.  We provide  the actual
  numbers in the supplementary material.}
\label{fig:towncentre}
\vspace{-0.25cm}
\end{figure}

%% file: figures/barplot.tex
\begin{table}[!h]
  \setlength\tabcolsep{4.5pt}
  \begin{tabular}{|l|c|c|c|c|}
\hline
    Approach & $\Delta\text{\IDF{}}$ & $\Delta\text{\IDF{}}^u$ & $\Delta\text{\MOTA{}}$ & $\Delta\text{\MOTA{}}^u$ \\
\hline
\KSP{} & 0.16 & 0.15 & -0.01 & -0.01 \\
\MDP{} & 0.05 & 0.02 & 0.03 & -0.01 \\
\RNN{} & 0.04 & 0.03 & 0.00 & -0.02 \\
\SORT{} & 0.04 & 0.02 & 0.06 & 0.00 \\
\hline
\end{tabular}
\caption{Improvement in \IDF{}  and  \MOTA{} metrics delivered  by  our  approach,
  averaged  over  all datasets.  The  2nd  and  4th  columns correspond  to  the
  supervised case, the 3rd and 5th to the unsupervised one.}
\label{fig:barplot}
\vspace{-0.4cm}
\end{table}

\comment{
\begin{table}[!h]
  \begin{tabular}{|l|c|c|c|c|}
\hline
    Approach & $\Delta\text{\IDF{}}^*$ & $\Delta\text{\IDF{}}^i$ & $\Delta\text{\MOTA{}}^*$ & $\Delta\text{\MOTA{}}^i$ \\
\hline
\Hotel{} & 0.05 & 0.04 & 0.05 & -0.02 \\
\ETH{} & 0.06 & 0.03 & 0.01 & 0.01 \\
\Station{} & 0.03 & 0.04 & -0.01 & -0.02 \\
\PTZ{} & 0.01 & -0.40 & 0.02 & -0.82 \\
\PTZ{}-f & -- & 0.02 & -- & 0.00 \\
\hline
\end{tabular}
\caption{Iprovement  by our  approach  for various  datasets,  averaged over  4
methods, \KSP{}, \MDP{}, \RNN{}, and \SORT{}.  Small number of people and large
number of patterns makes the method fail  on \PTZ{} dataset. However, if we use
iterative approach on the whole dataset,  denoted \PTZ{}-f, instead of 1 minute
test  segments, we  see  small but  persistent improvement  in  \IDF{} for  all
methods. In such setting, supervised approach is not possible due to absence of
training data.}
\label{fig:barplot}
\vspace{-0.25cm}
\end{table}
}

\comment{
\begin{figure}[!h]
\hspace{-1pt}
  \hspace{-0.4cm}\includegraphics[width=\columnwidth]{drawings/barplot.png}
\caption{Barplot}
\label{fig:barplot}
\end{figure}
}

%% file: figures/pictures.tex
\begin{figure*}[!t]
\begin{center}
\hspace{-1pt}
\begin{tabular}{ccccc}
  \hspace{-0.4cm}\includegraphics[trim={0cm 0cm 0cm 0cm},clip,width=0.20\textwidth]{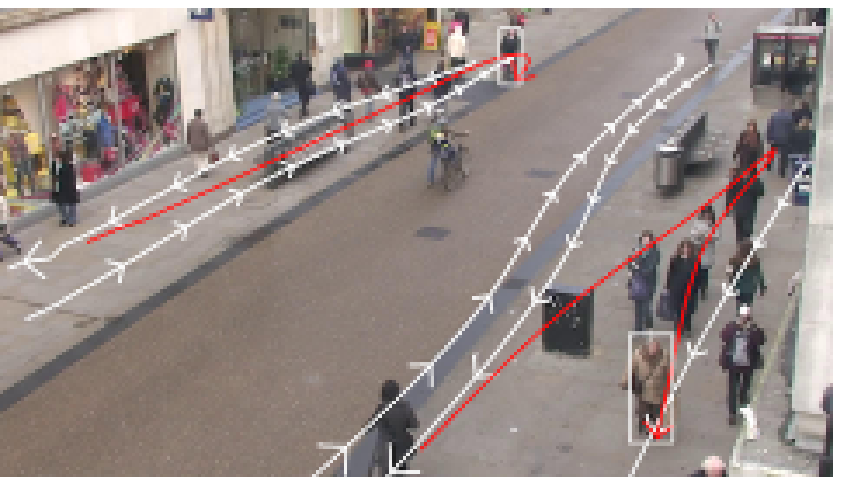} &
  \hspace{-0.4cm}\includegraphics[trim={0cm 0cm 0cm 0cm},clip,width=0.20\textwidth]{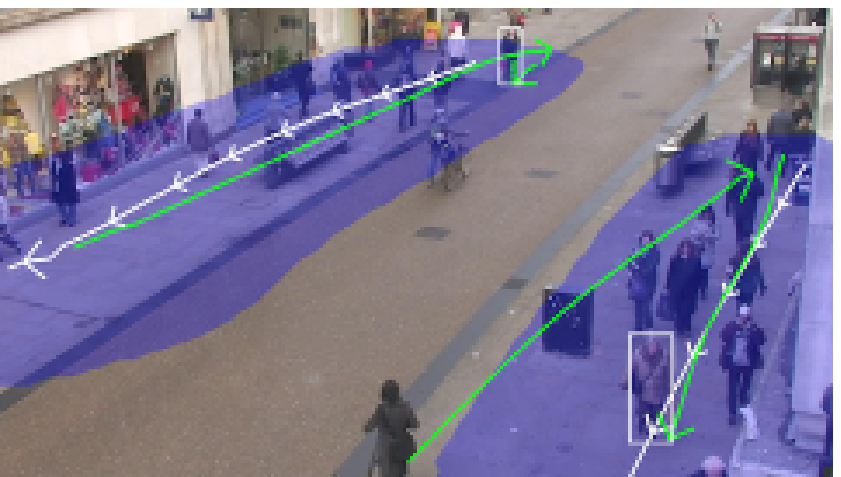} &
  \hspace{-0.4cm}\includegraphics[trim={0cm 0cm 0cm 0cm},clip,width=0.20\textwidth]{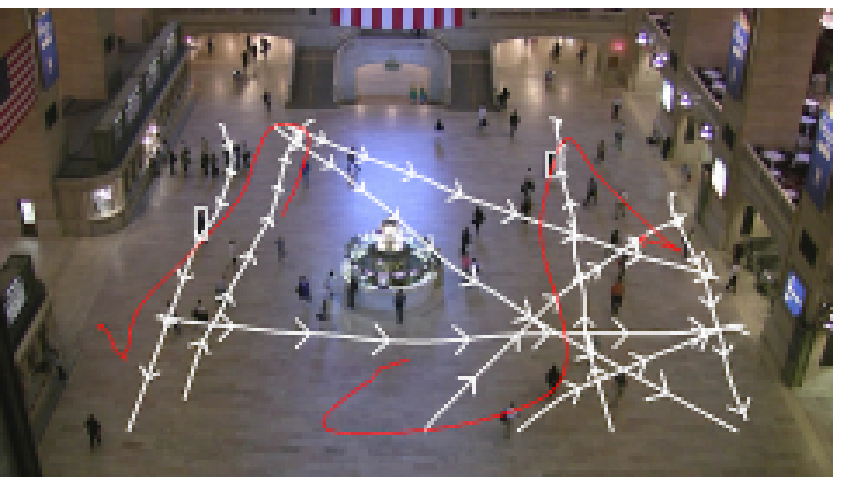} &
   \hspace{-0.4cm}\includegraphics[trim={0cm 0cm 0cm 0cm},clip,width=0.20\textwidth]{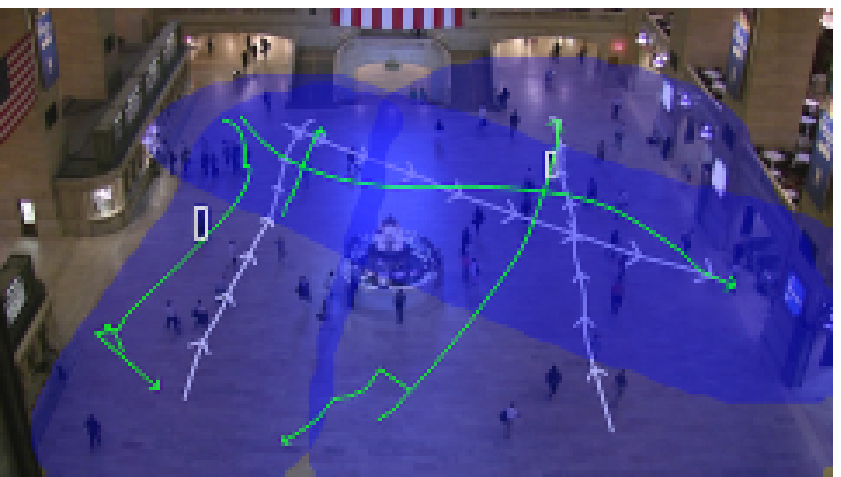}
&
   \hspace{-0.4cm}\includegraphics[trim={0cm 0cm 0cm 2.15cm},clip,width=0.20\textwidth]{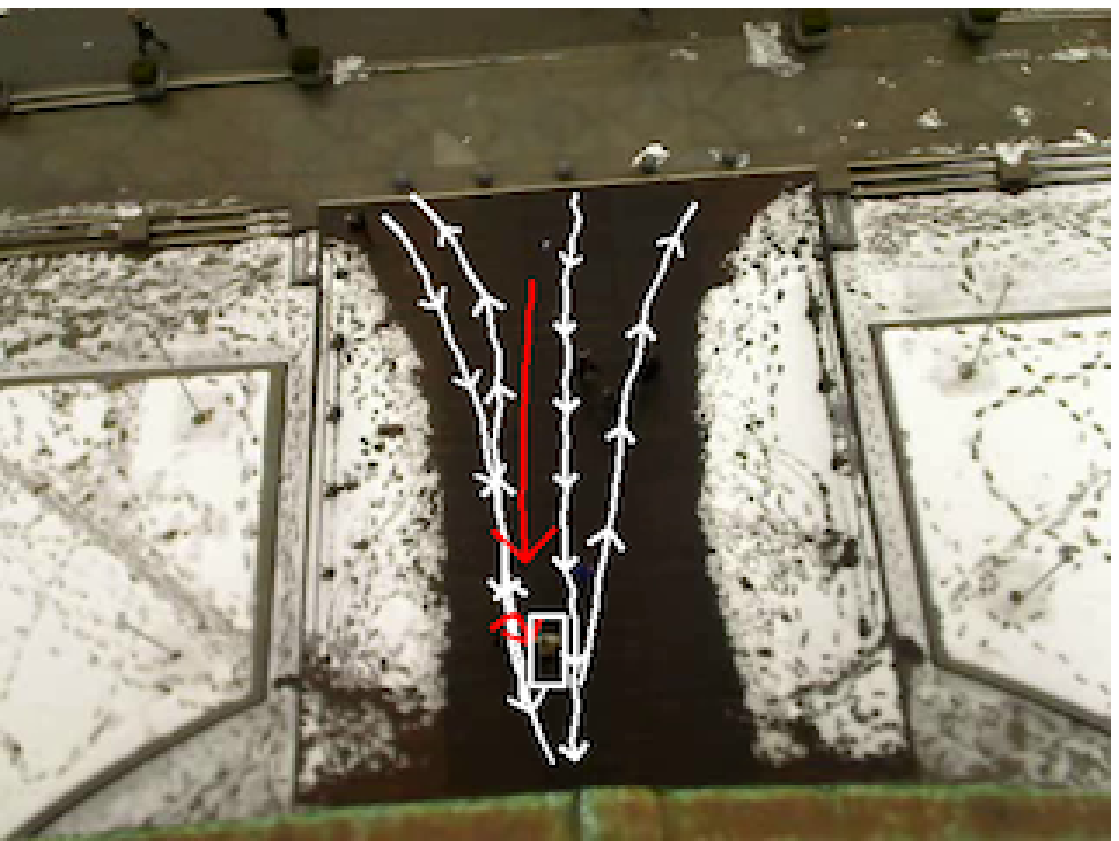} \\
   \hspace{-0.2cm} (a) &
   \hspace{-0.2cm} (b) &
   \hspace{-0.2cm} (c) &
   \hspace{-0.2cm} (d) &
   \hspace{-0.2cm} (e) \\
\end{tabular}
\end{center}
\vspace{-0.4cm}
\caption{Examples  of   learned  patterns,  denoted  by   their  centerline  in
white,  with some  erroneous trajectories  found by  various baselines  in red.
White  bounding  boxes  for  people   following  the  trajectories  are  shown.
Improved  trajectories  found by  our  approach  in  green.  We also  show  the
pattern  widths  (area in  blue),  to  show that  the  trajectory  we found  is
assigned  to  a  particular  pattern. {\bf  (a)}  \Towncentre{}  dataset,  {\bf
EAMTT~\cite{Sanchez-Matilla16}}  merges   several  trajectories  going   in  in
opposite  directions,  but  {\bf  (b)}  correct  pattern  assignment  helps  to
fix  that; {\bf  (c)}  Using  only affinity  information,  \KSP{}  is prone  to
multiple identity switches; {\bf (d)} Our approach recovers several trajectories
correctly, but  merges trajectories of two  different people in the  lower left
corner going in the same general direction; {\bf (e)} \ETH{} dataset, due to low
visibility using flow and feature point  tracking is hard, and \MDP{} fragments
a single trajectory into two, but our approach fixes that (not shown)}
\label{fig:pictures}
\end{figure*}

%% file: figures/iterative.tex
\begin{figure*}[!t]
\begin{center}
\hspace{-1pt}
\begin{tabular}{ccccc}
  \hspace{-0.4cm}\includegraphics[trim={0cm 0cm 0cm 0cm},clip,width=0.20\textwidth]{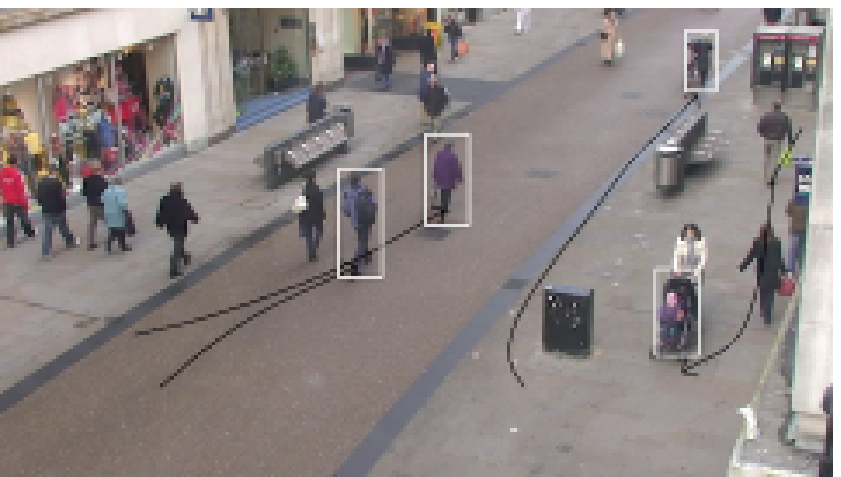} &
   \hspace{-0.4cm}\includegraphics[trim={0cm 0cm 0cm 0cm},clip,width=0.20\textwidth]{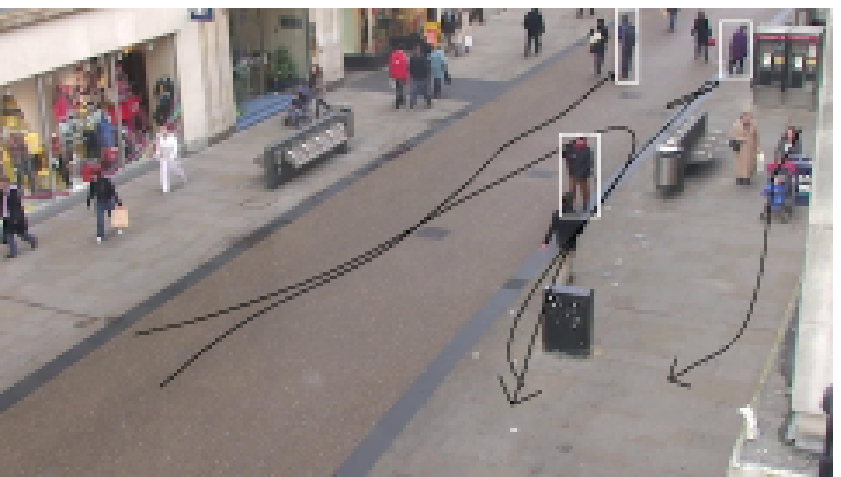} &
   \hspace{-0.4cm}\includegraphics[trim={0cm 0cm 0cm 0cm},clip,width=0.20\textwidth]{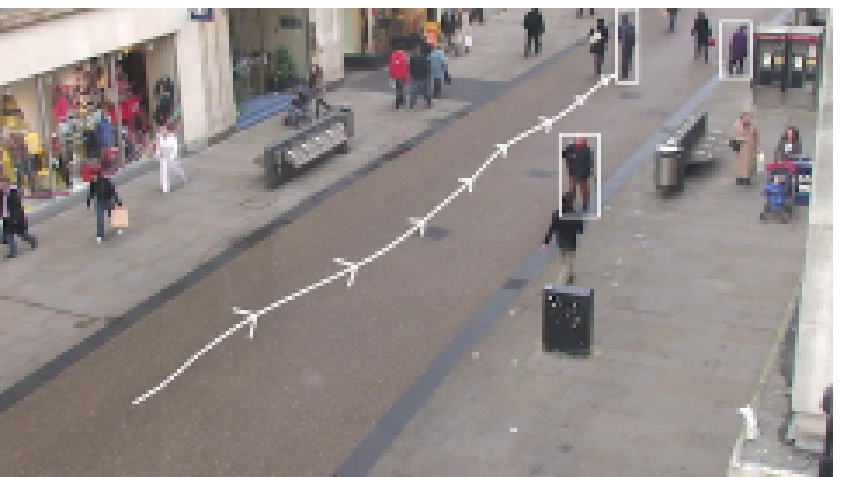} &
   \hspace{-0.4cm}\includegraphics[trim={0cm 0cm 0cm 0cm},clip,width=0.20\textwidth]{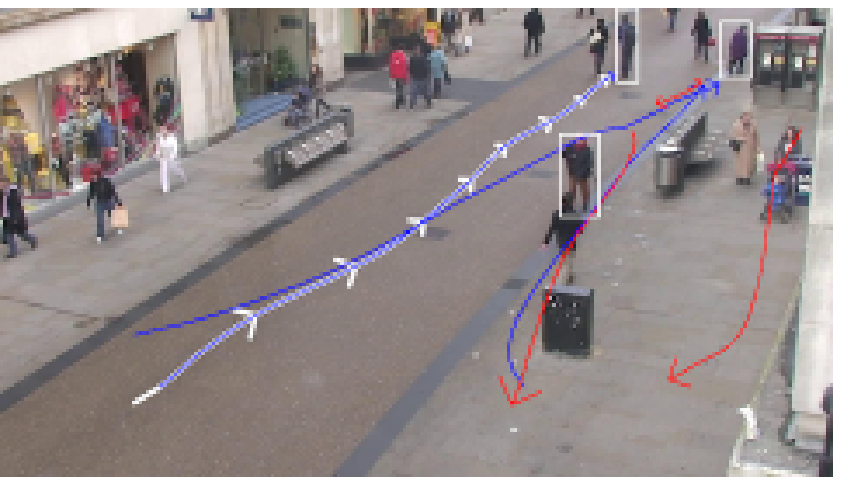} &
   \hspace{-0.4cm}\includegraphics[trim={0cm 0cm 0cm 0cm},clip,width=0.20\textwidth]{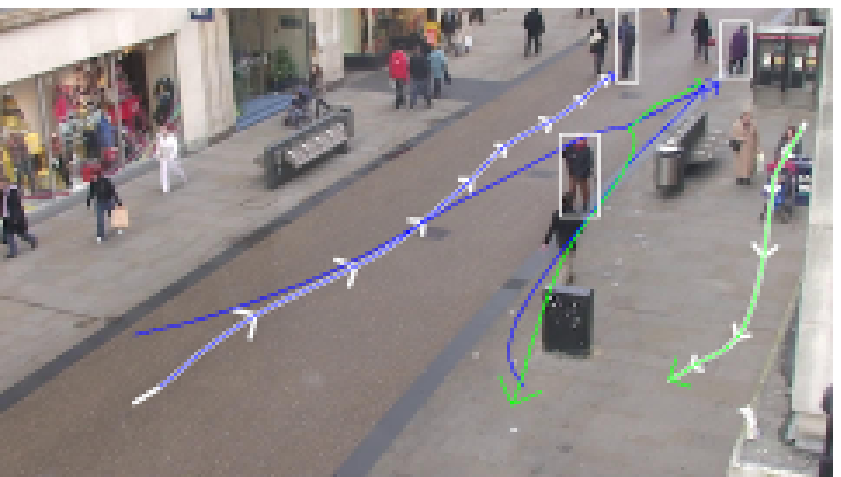} \\
   \hspace{-0.2cm} (a) &
   \hspace{-0.2cm} (b) &
   \hspace{-0.2cm} (c) &
   \hspace{-0.2cm} (d) &
   \hspace{-0.2cm} (e) \\
\end{tabular}
\end{center}
\vspace{-0.4cm}
\caption{Example  of  unsupervised  optimization.  {\bf (a)  }Four  people  are
tracked using  \KSP{}. Trajectories  are shown as  solid black  lines, bounding
boxes  are  white.  {\bf  (b)  }Tracks  continue,  featuring  several  identity
switches.  {\bf (c)  }First  step  of the  alternating  scheme  finds a  single
pattern, in  white, that explains as  many trajectories as possible,  it is the
leftmost  trajectory.  {\bf (d)  }Given  this  pattern,  next  step is  to  fit
trajectories to  it. Trajectories in  blue are the ones assigned to  this pattern,
trajectories in red  are assigned to no pattern. One  identity switch is fixed.
{\bf  (e) }  After  several iterations,  we  look for  the  best two  patterns.
Rightmost trajectory is  picked as the second pattern.  Fitting trajectories to
the  best two  patterns  allows  to fix  the  remaining fragmented  trajectory.
Trajectories assigned to the second pattern in green.}
\label{fig:iterative}
\vspace{-0.3cm}
\end{figure*}

%% file: figures/eval-gt-idf.tex
\begin{table}[!h]
  \begin{tabular}{|p{1.6cm}|c|c|c|c|c|}
\hline
Approach: Dataset: & \MDP{} & \RNN{} & \SORT{}  & \KSP{} & \OUR{} \\
\hline
\Towncentre{}      & 0.87   & 0.65   & 0.88     & 0.55   & \bf 0.93 \\
\hline
\ETH{}             & 0.89   & 0.65   & \bf 0.93 & 0.59   & 0.92 \\
\hline
\Hotel{}           & 0.85   & 0.70   & 0.88     & 0.60   & \bf 0.94 \\
\hline
\Station{}         & 0.68   & 0.40    & \bf 0.72     & 0.45  & 0.70 \\
\hline
\end{tabular}
\caption{\IDF{} evaluation results using ground truth detections.}
\label{tab:eval-gt-idf}
\vspace{-0.5cm}
\end{table}

%% file: figures/eval-gt-mota.tex
\begin{table}[!h]
  \begin{tabular}{|p{1.6cm}|c|c|c|c|c|}
\hline
Approach: Dataset: & \MDP{} & \RNN{} & \SORT{}  & \KSP{}   & \OUR{} \\
\hline
\Towncentre{}      & 0.87   & 0.85   & 0.90     & 0.87     & \bf 0.98 \\
\hline
\ETH{}             & 0.85   & 0.73  & 0.85     & 0.70     & \bf 0.94 \\
\hline
\Hotel{}           & 0.84   & 0.78   & 0.82     & 0.74     & \bf 0.97 \\
\hline
\Station{}         & 0.75    & 0.68   & 0.70     & \bf 0.80     & 0.77 \\
\hline
\end{tabular}
\caption{\MOTA{} evaluation results using ground detections.}
\label{tab:eval-gt-mota}
\end{table}

%% file: figures/time.tex
\begin{table}[!h]
  \setlength\tabcolsep{5pt}
  \begin{tabular}{|l|c|c|c|c|c|}
\hline
Dataset      & \Towncentre{} & \ETH{} & \Hotel{} & \Station{} & \Station{} \\ \hline
Frames       & 150           & 227    & 268      & 75         & 75 \\ \hline
Trajectories & 85            & 67     & 47       & 100        & 193 \\ \hline
Patterns     & 7             & 5      & 4        & 26         & 26 \\ \hline
Detections   & 2487          & 894    & 1019     & 1960       & 3724 \\ \hline
Variables    & 70k           & 17k    & 18k      & 191k       & 450k \\ \hline
Time, s      & 26            & 4      & 4        & 160        & $>$3600 \\ \hline
\end{tabular}
\caption{Optimization problem size and run time of our approach for processing
  a typical one mn batch from each dataset.}
\label{tab:time}
\vspace{-0.5cm}
\end{table}

%% file: conclusion.tex

\section{Conclusion}
\label{sec:conclusion}

In this  work we  have proposed  an approach to  tracking multiple  people under
global behavioral constraints.  It lets us learn motion patterns given
ground  truth trajectories, use  these patterns  to guide the  tracking, and
improve  upon a  wide range  of state-of-the-art  approaches. It also
extends naturally to the unsupervised case without ground truth.

Our optimization  scheme is generic and  allows for a wide  range of definitions
for the patterns, beyond the ones we have  used here.  In the future, we plan to
to work  with more complex patterns  that models human behavior  better, account
for appearance, and handle correlations between people's behavior.

%% file: supplementary.tex
In  Section~\ref{sec:mn},  we  provide  the full  definitions  of  the  scoring
functions   $n$  and   $m$,  described   in  Section~3.3   of  the   paper.  In
Section~\ref{sec:opt}, we provide additional  details of the optimizations used
to improve the output  of other method and to learn  the patterns, described in
Sections~4.1 and  4.2 of the  paper. In Section~\ref{sec:res}, we  provide full
results of all  methods on all datasets with various  metrics, extending on the
results from Section~6.4  of the paper. Finally,  in Section~\ref{sec:time}, we
provide  evaluation  of the  computational  requirements  of our  approach,  in
addition to the results given in Secton~6.4 of the paper.
\clearpage

\section{Full definitions of $n$ and $m$ functions}
\label{sec:mn}

These functions  are used  to score the  edges of a  trajectory to  compute how
likely is  it that  a particular  trajectory follows  a particular  pattern. As
stated in Section 3.3 of the paper:

\begin{small}
\begin{eqnarray}
  C(T,P,A) \mbox{\hspace{-0.1cm}} & = \mbox{\hspace{-0.1cm}} &
  {\sum\limits_{t \in T} M(t,p_{A(t_1)})
  \over
  \sum\limits_{t \in T} N(t,p_{A(t_1)})}, \label{eq:costFun} \\
  N(t, p) \mbox{\hspace{-0.1cm}} & = \mbox{\hspace{-0.1cm}}& n(I, t_1, p) + n(t_{|t|}, O, p) + \mbox{\hspace{-0.3cm}} \sum\limits_{
    1 \leq j \leq |t| - 1} \mbox{\hspace{-0.3cm}} n(t_j, t_{j + 1}, p), \\
  M(t, p) \mbox{\hspace{-0.1cm}} & = \mbox{\hspace{-0.1cm}}& m(I, t_1, p) + m(t_{|t|}, O, p) + \mbox{\hspace{-0.5cm}} \sum\limits_{
    1 \leq j \leq |t| - 1} \mbox{\hspace{-0.3cm}}m(t_j, t_{j + 1}, p),
  \end{eqnarray}
\end{small}\noindent
where $T$ is a set of edges  of all trajectories, $A$ is the assignment between
a trajectory and a  pattern, and $P$ is a set of patterns.  As shown in (2) and
(3), to score a trajectory we score all  its edges plus the edges from $I$, the
node denoting the beginnings of the trajectories, and the ones to $O$, the node
denoting the ends of trajectories. As  mentioned in the paper, we want $N(t,p)$
to reflect the full  length of the trajectory and the  pattern, and $M(t,p)$ to
reflect the  total length of  the aligned trajectory  and the pattern.  In what
follows, we provide definitions of $n$ and $m$ in all cases.

\begin{figure}[h]
\begin{center}
\hspace{-1pt}
\includegraphics[trim={11cm 8cm 10cm 9cm},clip,width=0.5\textwidth]{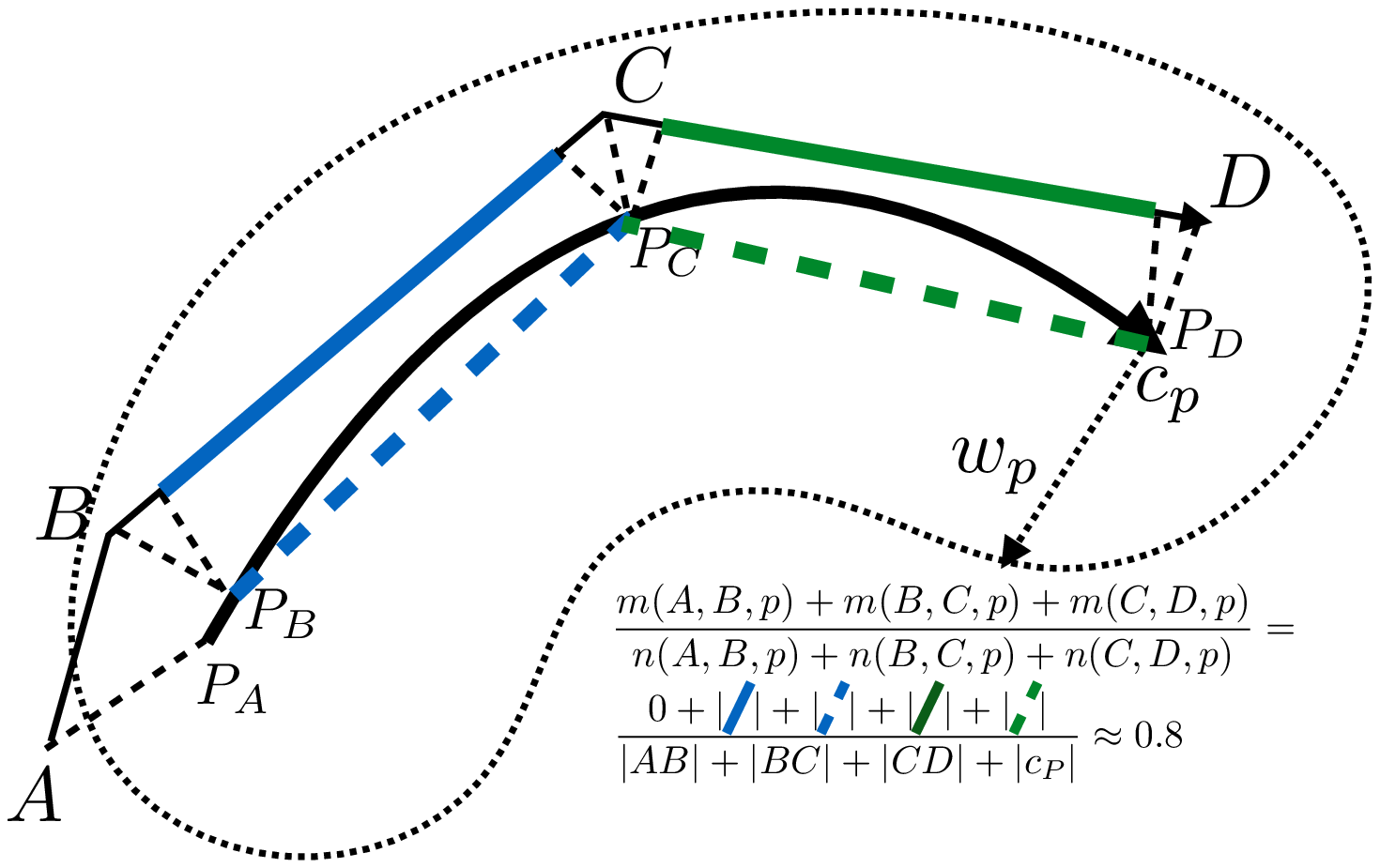}
\end{center}
\begin{small}
	\vspace{-1cm}
\caption{Example of computing the cost function $C$ for three consecutive edges
$(A,B)$,  $(B,C)$, $(C,D)$.  Dotted line  around the  pattern centerline  $c_p$
shows  the area  within  the distance  $w_p$ to  the  pattern. The  denominator
contains the total  length of the edges  plus the total length  of the pattern,
while the  numerator contains the parts  aligned with each other  (in green and
blue). The edge $(A,B)$ is not counted  as aligned, because $A$ is further from
the pattern than its width $w_p$.}
\end{small}
\label{fig:n-and-m-functions}
\end{figure}

\noindent
In {\bf  Table~\ref{tab:nm-normal}}, we  show how  to compute  $n$ and  $m$ for
edges that  link two detections  and follow some pattern.  For $n$ we  take the
pattern length to  be positive or negative depending on  whether the projection
of the edge to the pattern is  positive or negative. For $m$, we penalize edges
far from the pattern and edges going  in the direction opposite to the pattern,
in two different ways, which gives rise to the three cases shown in the table.
\noindent
In {\bf  Table~\ref{tab:nm-io}}, we  show how  to compute $n$  when one  of the
nodes is $I$ or  $O$, denoting the start or the end of  a trajectory. A special
case arises when a node is in the first or the last frame of an input batch, and
a trajectory going  through it does not need to  follow the pattern completely.
This results in a total of two cases we show in the table.
\noindent
In  {\bf Table~\ref{tab:nm-nop}},  we show  the two  cases when  we assign  the
transition to  no pattern $\emptyset$,  one case when  we assign a  normal edge
joining two detections, and  the other when we assign edge from  $I$ or to $O$,
indicating the beginning or the edge of the trajectory.

\begin{table}[!t]
\begin{tabular}{|p{0.2\textwidth}|p{0.25\textwidth}|c|}
  \hline
  {\bf Case} & {\bf Explanation} & {\bf Figure} \\
  \hline
  Normal edge aligned with the pattern: $B$ and $C$ are within distance $w_p$ to
  the pattern centerline, $P_B$ is  earlier on the curve $c_p$ that $P_C$. &
  For the edge $(B,C)$, we find the nearest neighbor of the two endpoints on
  the pattern, namely $P_B$ and $P_C$.
  Formally, we have $P_B=\arg\min\limits_{x \in c_p}||B-x||$. Then we project $P_B$ and $P_C$ orthogonally back onto $(B,C)$. This guarantees that $m(B, C, p) \le n(B, C, p)$ with equality when $(B, C)$ and $(P_B, P_C)$ are two parallel segments of equal length, and also penalizes deviations from the pattern in direction. &
  \includegraphics[valign=T,trim={11cm 8cm 10cm 8cm},clip,width=0.45\textwidth]{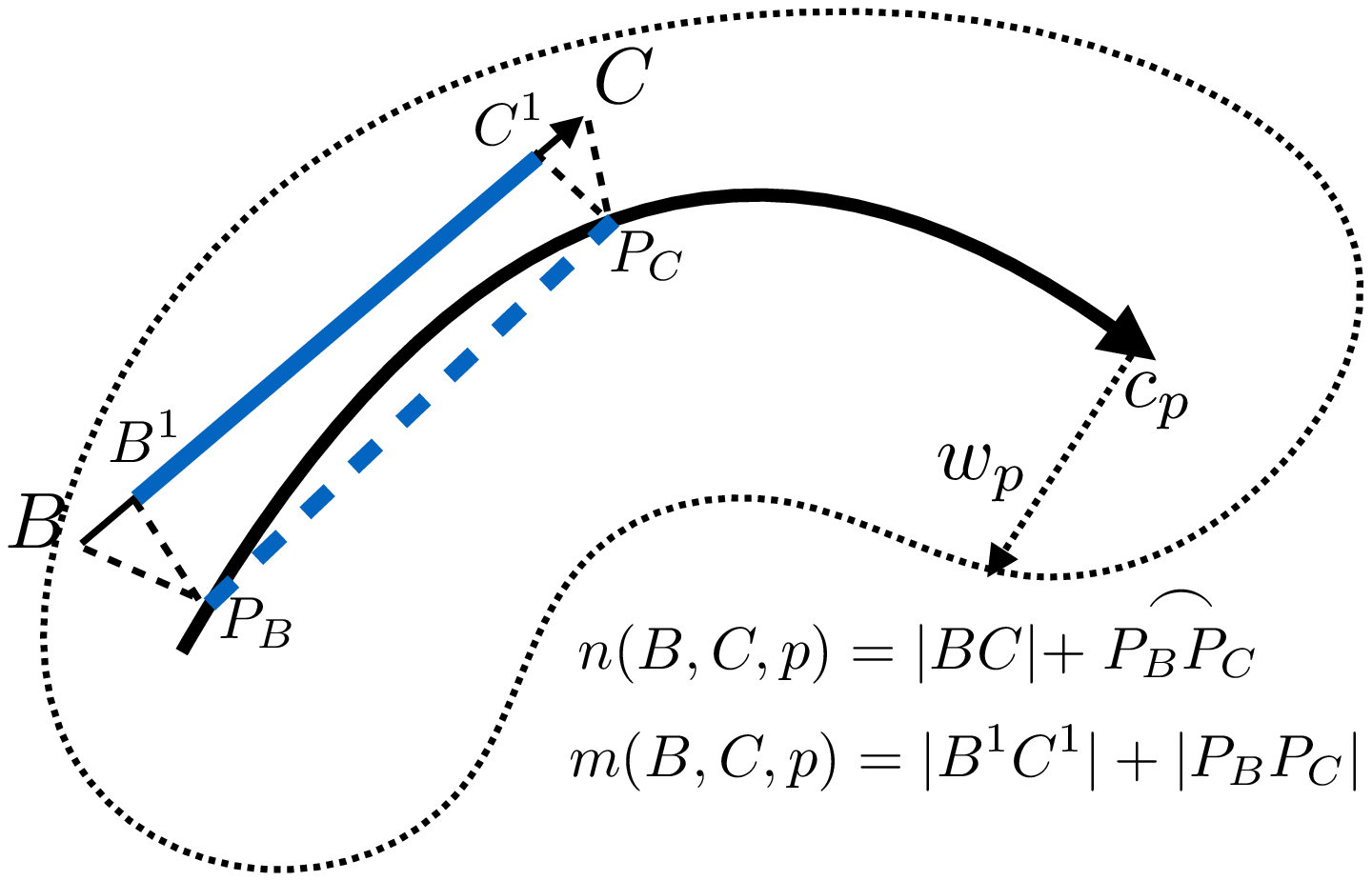} \\
  \hline

  Normal edge aligned with the pattern: $B$ and $C$ are further away than $w_p$ from the pattern centerline, $P_B$ is earlier on the curve $c_p$ that $P_C$. &
  $n(B, C, p)$ is calculated in the same way as done in the previous case. To penalize deviations from the pattern in distance, we take $m(B, C, p) = 0$ &
  \includegraphics[valign=T,trim={10cm 8cm 10cm 7cm},clip,width=0.45\textwidth]{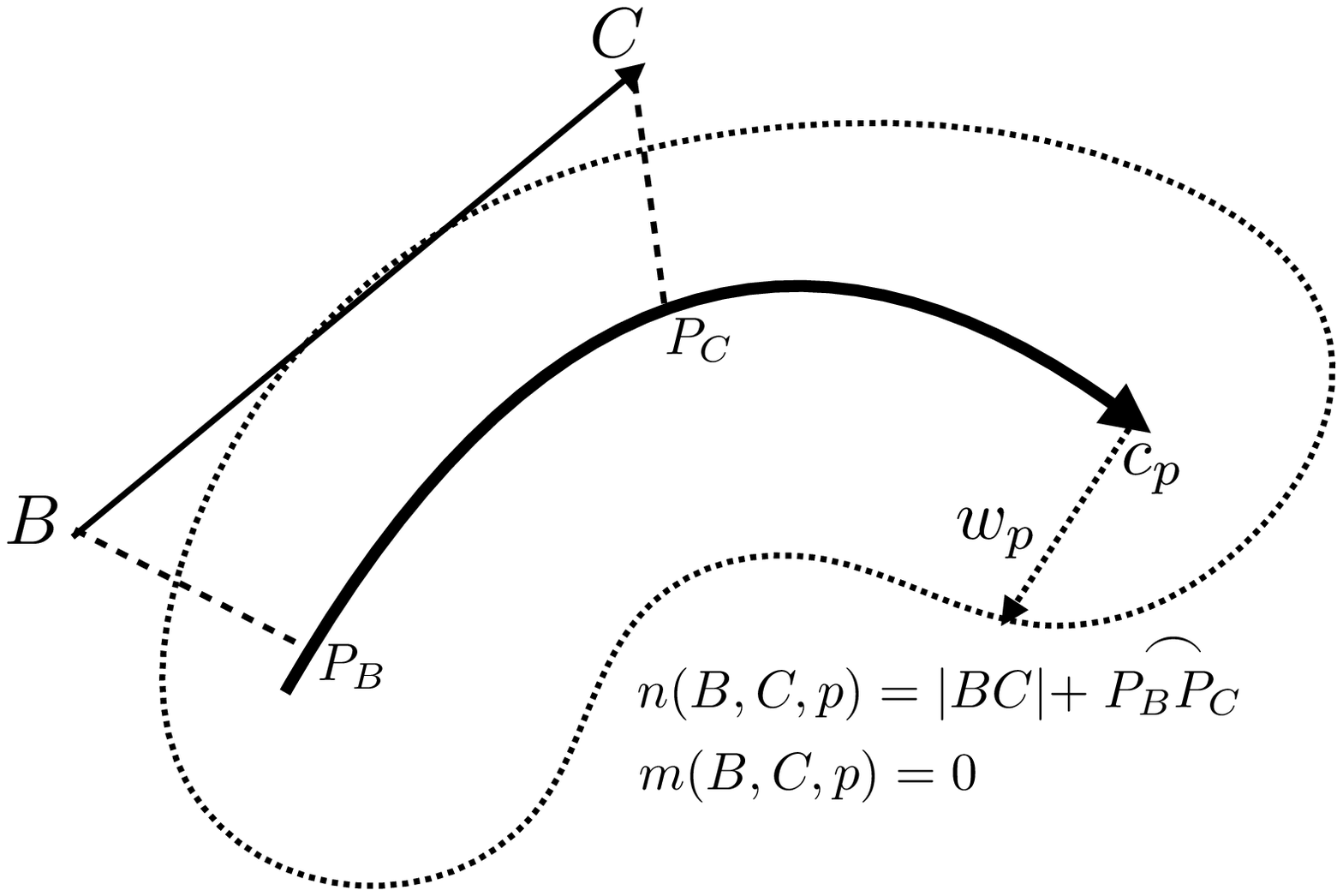} \\
  \hline

  Normal edge not aligned with the pattern: $P_B$ is later on the curve $c_p$ that $P_C$. &
  To keep our rule about $N$ being the sum of lengths of pattern and trajectory, we need to subtract the length of arc from $P_B$ to $P_C$, as it is pointing in the direction opposite to the pattern. To penalize this behavior, we take $m(B, C, p)$ to be $-|P_B P_C|$, multiplied by $1 + \epsilon$. In practice, we use $\epsilon=1$. &
  \includegraphics[valign=T,trim={11cm 8cm 9cm 8cm},clip,width=0.45\textwidth]{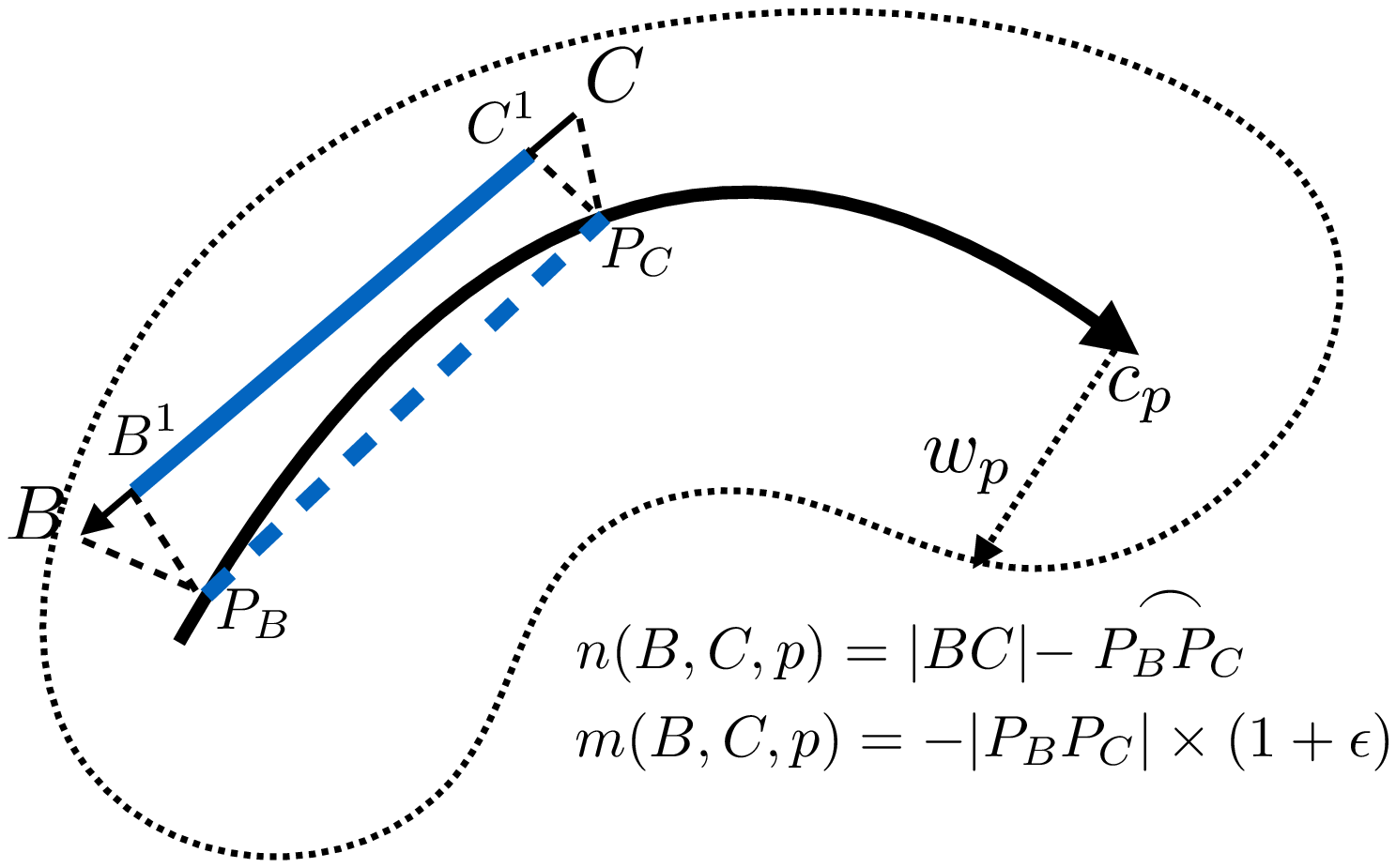} \\
  \hline
\end{tabular}
\caption{Table describing full definitions of $n$ and $m$ in normal cases, when edges between two detections align with a pattern. They all follow naturally from the rule about $N$ being the sum of length of trajectory and the pattern, and $M$ being the sum of aligned lengths.}
\label{tab:nm-normal}
\end{table}
\clearpage

\begin{table}[!t]
\begin{tabular}{|p{0.2\textwidth}|p{0.25\textwidth}|c|}
  \hline
  {\bf Case} & {\bf Explanation} & {\bf Figure} \\ \hline

  Edge from the source to a normal node / from a normal node to the sink&
  To keep our rule about $N$ being the sum of lengths of pattern and trajectory, we need to add the length from the beginning of the pattern to the point closest to the node on the centerline / from the point closest to the node on the centerline to the end of the pattern. Since we didn't observe any parts of trajectory aligned with these parts, we take $m=0$. &
  \includegraphics[valign=T,trim={6cm 8cm 8cm 5cm},clip,width=0.45\textwidth]{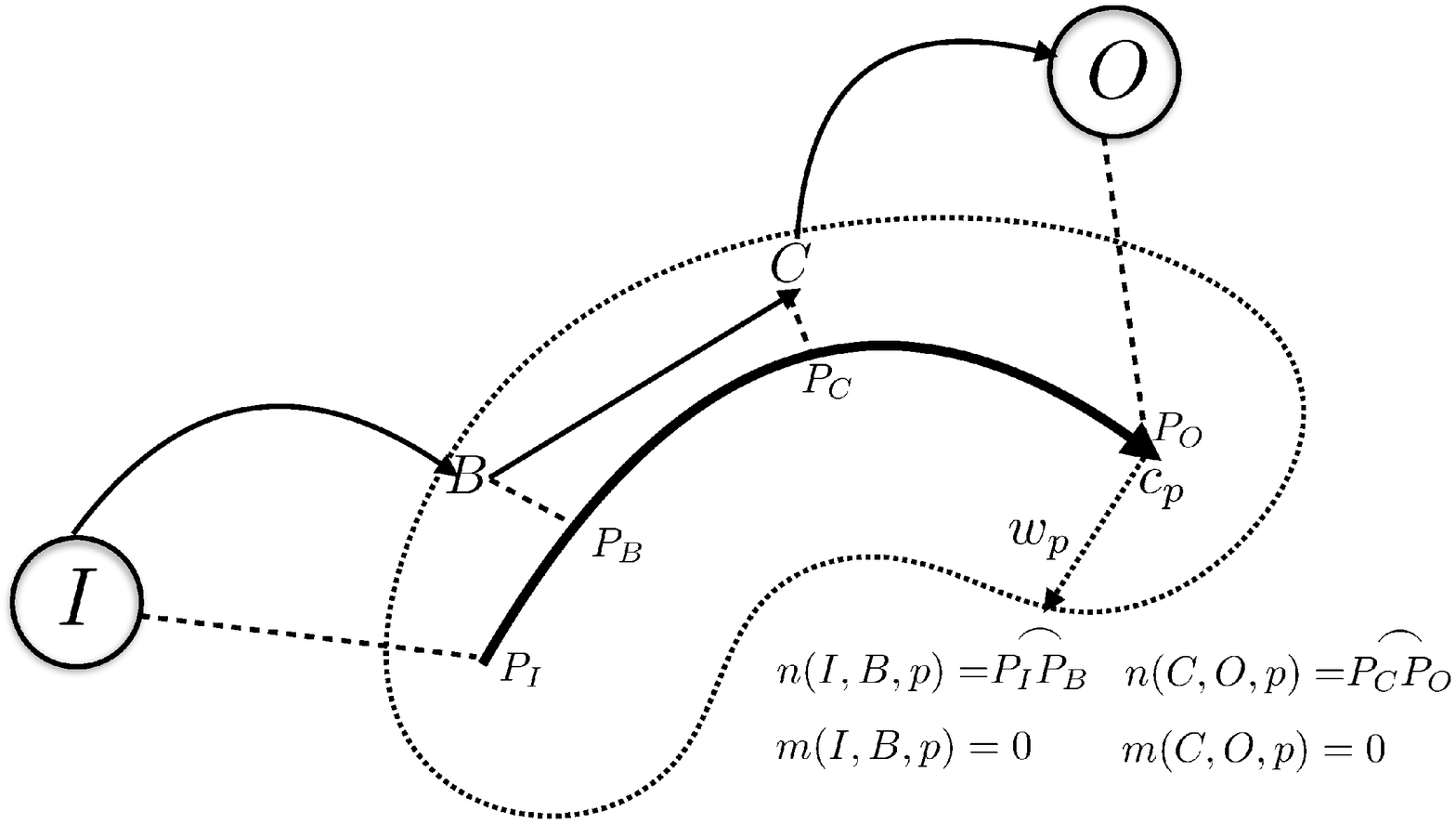} \\
  \hline

  Edge from the source to a normal node in the first frame of the batch / from a normal node in the last frame of the batch to the sink &
  We assume that our trajectories  follow the path completely. However, this might be not true, which we observe from the middle, that is, the ones that begin in the first frame of the batch or end in the last frame. In that case we don't need to add the part of the pattern before / after the current point closest to the node, which is why we take  $n=m=0$. &
  \includegraphics[valign=T,trim={6cm 8cm 10cm 5cm},clip,width=0.45\textwidth]{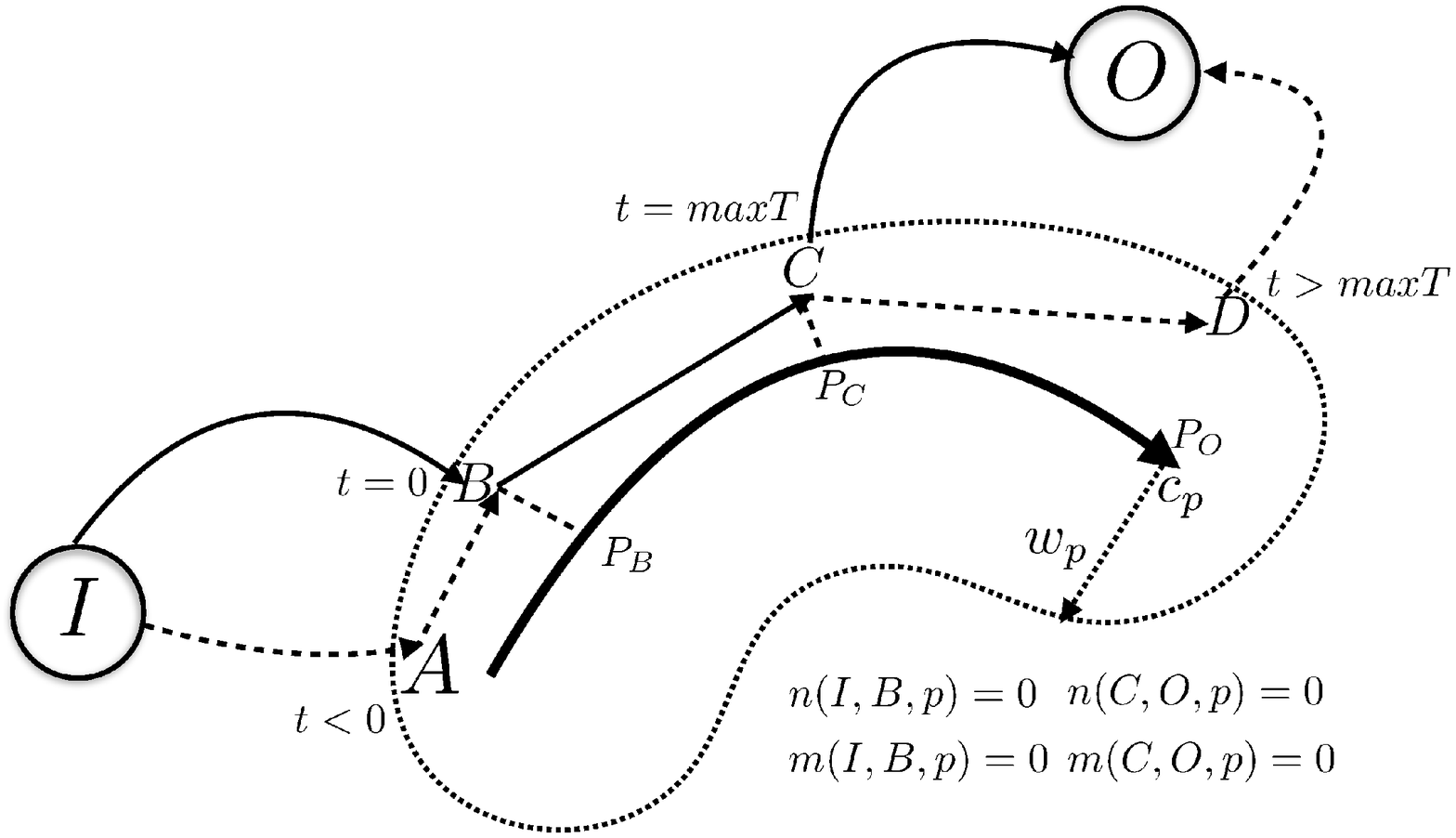} \\
  \hline
\end{tabular}
\caption{Table describing full definitions of $n$ and $m$ in corner cases when one of the edges go through $I$ or $O$, indicating the beginning or the end of a trajectory. They all follow naturally from the rule about $N$ being the sum of length of trajectory and the pattern, and $M$ being the sum of aligned lengths.}
\label{tab:nm-io}
\end{table}
\clearpage

\begin{table}[!t]
\begin{tabular}{|p{0.2\textwidth}|p{0.25\textwidth}|c|}
  \hline
  {\bf Case} & {\bf Explanation} & {\bf Figure} \\ \hline
  Normal edge aligned to no pattern &
  To keep our rule about $N$ being the sum of lengths, we take $n$ to be just the length of the trajectory, since we assume the length of empty pattern to be zero. We penalize such assignment by a fixed constant $\epsilon_{\emptyset}$, taking $m$ to be $n$ multiplied by such constant. In practice, we keep $\epsilon_{\emptyset}=0.3$ when training from ground truth, or $\epsilon_{\emptyset}=-3$ otherwise. &
  \includegraphics[valign=T,trim={10cm 8cm 10cm 6cm},clip,width=0.45\textwidth]{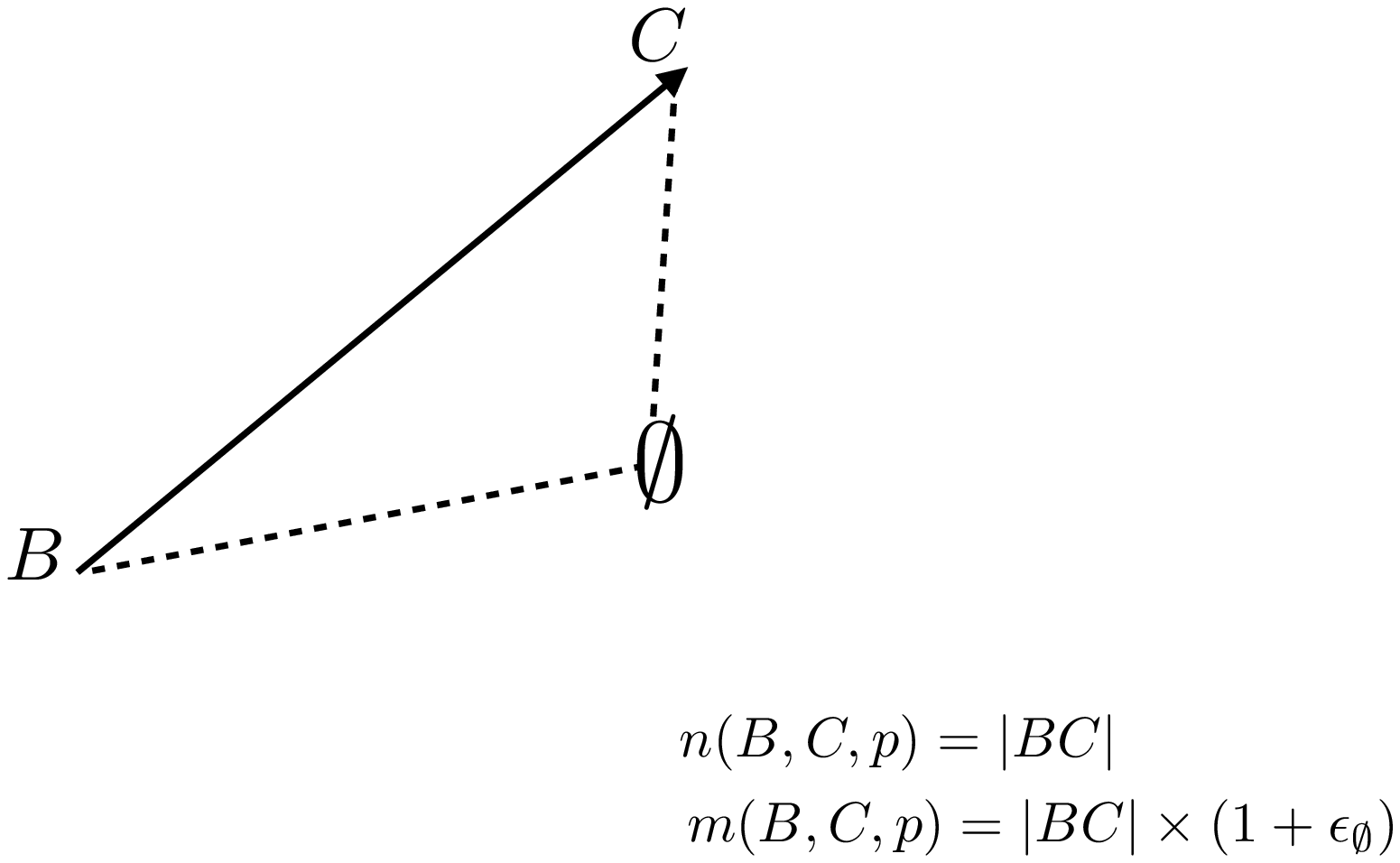} \\
  \hline

  Edge from the source / to the sink, aligned to no pattern &
  To keep our rule about $N$, we take both $n=m=0$. &
  \includegraphics[valign=T,trim={6cm 8cm 10cm 7cm},clip,width=0.45\textwidth]{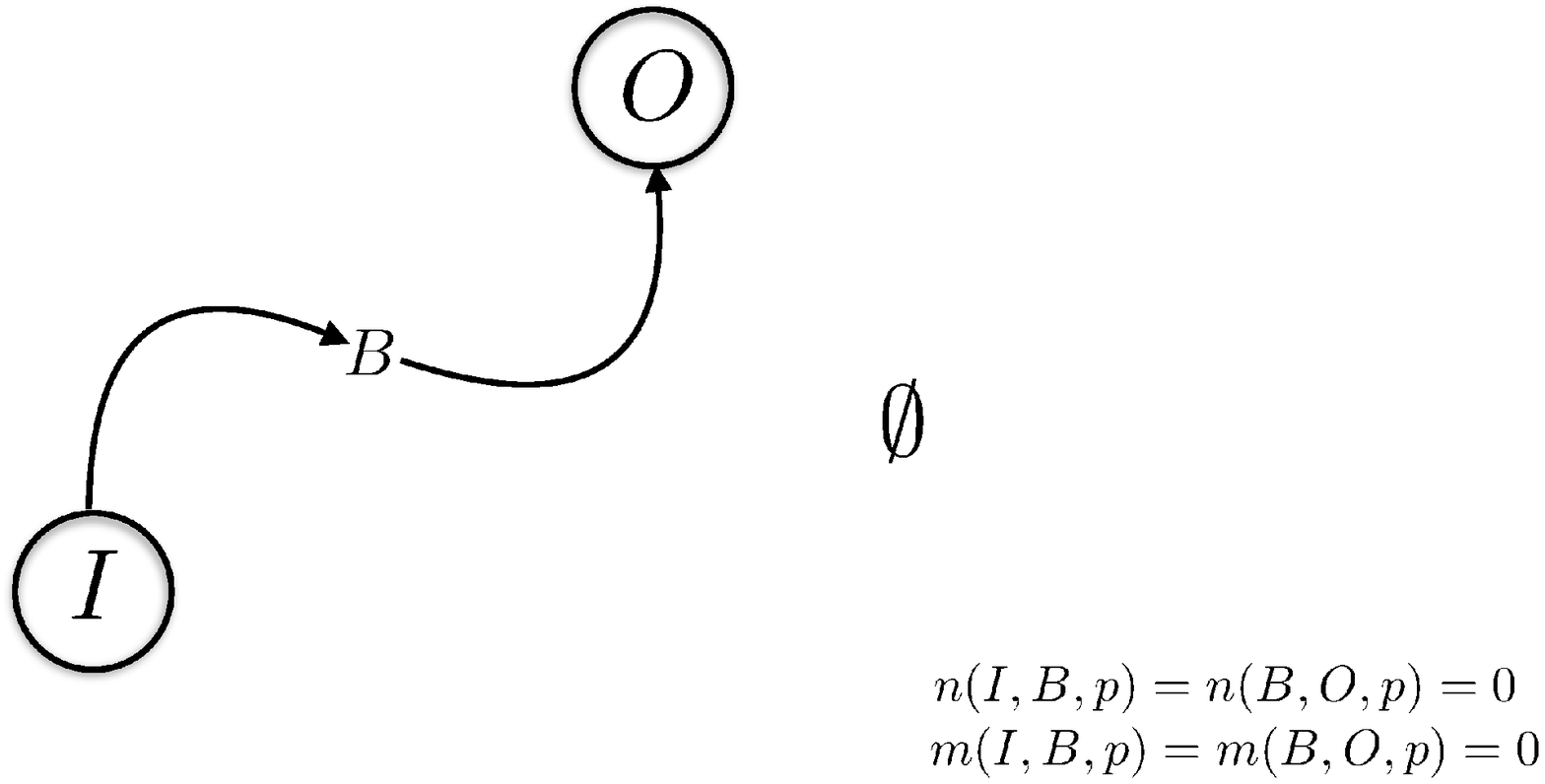} \\
  \hline
\end{tabular}
\caption{Table describing full definitions of $n$ and $m$ in corner cases when there is no pattern.
They all follow naturally from the rule about $N$ being the sum of length of trajectory and the pattern,
and $M$ being the sum of aligned lengths.}
\label{tab:nm-nop}
\end{table}
\clearpage

\newpage

\section{Details of the optimization scheme}
\label{sec:opt}

Here we provide  details on our optimization schemes that  improve the tracking
output of other method and learn patterns,  outlined in Sections 4.1 and 4.2 of
the paper, respectively.

\subsection{Tracking}

As noted in the paper, we  introduce the binary variables $o^p_{ij}$, denoting
the  number  of  people  transitioning  between the  detections  $i$  and  $j$,
following pattern $p$. We put the following constraints on them:

\vspace{-0.1cm}
\begin{eqnarray}
  \forall   i  \in   \mathcal{D}\cup{O}  \sum\limits_{(i,j)   \in  \mathcal{E},   p  \in
    P^*}o^p_{ij} & = & 1 \;, \label{eq:flowConstraints} \\
  \forall j \in \mathcal{D}, p \in  P^* \sum\limits_{(i,j) \in \mathcal{E}}o^p_{ij} &
  = & \sum\limits_{(j,k) \in \mathcal{E}}o^p_{jk} \;. \nonumber
\end{eqnarray}
Then, during  binary search, we fix  a particular value of  $\alpha$, and check
whether the problem constrained by~\eqref{eq:flowConstraints} and the following
has a feasible point:
\begin{equation}
  \label{eq:flowCostFun}
  \sum\limits_{(i, j) \in T, p \in P^*} (m(i,j,p) -\alpha n(i, j, p)) o^p_{ij} \ge 0
\end{equation}

If a feasible point  exists, we pick a value of $\alpha$ to  be the lower bound
of the best $\alpha$,  for which the problem is feasible,  otherwise we pick it
as an upper bound. We start with the upper bound of 1 and lower bound of 0, and
pick $\alpha$ as an average between  the upper and the lower bound (dichotomy).
We repeat  this process  10 times,  allowing us  to find  the correct  value of
$\alpha$ with the margin of $2^{-10}$.

\subsection{Patterns}

As noted in the paper, we  introduce the binary variables $a_{tp}$ denoting that a ground truth trajectory $t$ follows the pattern $p$, and binary variables $b_p$ denoting whether at least one trajectory follows the pattern $p$.

\vspace{-0.3cm}
\begin{eqnarray}
  a_{tp}&\in&    \{0,1\}
  \; , \forall t \in T^*, p \in \mathcal{P} \;, \nonumber\\
  b_{p}&\in&\{0,1\}\; ,  \forall p \in \mathcal{P} \; , \nonumber\\
  \sum\limits_{p    \in   \mathcal{P}}a_{tp}&   =&
  1 \;, \forall t \in T^* \; , \label{eq:matchConstraints}\\
  a_{tp}&  \le &  b_p \; , \forall t \in T^*, p \in
  \mathcal{P} \; .\nonumber
\end{eqnarray}

We then do the same binary search as described above to find the highest $\alpha$, for which there exists a feasible point to a set of constraints~\eqref{eq:matchConstraints} and the following:

\vspace{-0.1cm}
\begin{align}
\label{eq:matchFractionConstr}
& \sum\limits_{t \in T^*}\sum\limits_{p \in \mathcal{P}} (m(t, p) - \alpha n(t, p)) a_{tp} \ge 0 \;, \nonumber \\
& \sum\limits_{p \in \mathcal{P}}b_p \le \alpha_p \;,  \\
& \sum\limits_{p \in \mathcal{P}}b_p M(p) \le \alpha_c \;. \nonumber
\end{align}

We do five iterations of binary search, and we obtain the right value of $\alpha$ with
precision of $2^{-5}$.  To create a set of all  possible patterns $\mathcal{P}$
we combine  the set of all  possible trajectories in the  current batch (taking
only those that  start after the beginning  of the batch and  end before the
end of the  batch to make sure  they represent {\it full}  patterns of movement)
with a set of possible lengths. For  all datasets except {\bf Station}, our set
of possible lengths is \{0.5, 1, 3, 5,  7, 9, 11, 13, 15, 17\}, while for the
{\bf  Station} dataset  we  use \{0.05,  0.1,  0.2, 0.3,  0.4,  0.5\} of  the
tracking area,  since we don't  know the exact sizes of  the tracking area,  but only
estimated homography between the ground and image plane.

\newpage
\section{Full results}
\label{sec:res}

Here  we provide  the  full results  of all the methods on all the datasets.
Tables~\ref{tab:res1},~\ref{tab:res2} are the full versions of Table 2
of the paper, and Table~\ref{tab:res3} is the full version of Tables 3 and 4 of the
paper. In Tables~\ref{tab:res1},~\ref{tab:res2}, we compare the original output
of the  method with the improvements  brought by our approach  in both supervised and
unsupervised  manner. In  Table~\ref{tab:res3}, we  compare the  methods when
using the ground truth  set of detections as input. As in  the paper, we report
the results for the matching distances of 3m (0.1 of the tracking area for the {\bf
Station}  dataset), and  for  IDF\_1 metric  we  also show  results  for 1m  to
indicate that the ranking of the  methods does not change, but the improvement
brought by  our methods is  less visible due  to reconstruction errors  when we
estimate the  3D position  of the person  from the bounding  box. This  fact is
especially  highlighted by  the Table~\ref{tab:res3},  where difference  in the
metric computed for distances of 3m. and 1m. is especially large.

Specifically, We  report
the IDF\_1,  identity level  precision  and recall  IDPR  and  IDRC  defined
in~\cite{Ristani16}, as well as MOTA, precision  and recall PR and RC, and the
number of mostly  tracked MT, partially tracked PT and mostly  lost trajectories ML
defined in~\cite{Bernardin08}.


\begin{table}
  \setlength\tabcolsep{9.5pt}
\begin{tabular}{|l|l|c|c|c|c|c|c|c|c|c|}
  \hline
  Method    & Dataset     & IDF\_1 & IDPR & IDRC & MOTA & PR   & RC   & MT    & PT   & ML \\ \hline
EAMTT       & \bf Town    & 0.72 (0.59)  & 0.76 & 0.68 & 0.73 & 0.92 & 0.82 & 158   & 68   & 20 \\ \hline
EAMTT-i     & \bf Town    & 0.80 (0.63)  & 0.84 & 0.76 & 0.73 & 0.91 & 0.82 & 165   & 59   & 22 \\ \hline
EAMTT-o     & \bf Town    & 0.82 (0.65)  & 0.83 & 0.80 & 0.74 & 0.89 & 0.86 & 182   & 44   & 20 \\ \hline \hline
JointMC     & \bf Town    & 0.75 (0.63)  & 0.90 & 0.65 & 0.64 & 0.95 & 0.68 & 128   & 54   & 64 \\ \hline
JointMC-i   & \bf Town    & 0.77 (0.64)  & 0.91 & 0.66 & 0.64 & 0.95 & 0.68 & 129   & 52   & 65 \\ \hline
JointMC-o   & \bf Town    & 0.76 (0.62)  & 0.88 & 0.67 & 0.65 & 0.93 & 0.71 & 138   & 50   & 58 \\ \hline \hline
MHT\_DAM    & \bf Town    & 0.56 (0.45)  & 0.82 & 0.42 & 0.40 & 0.90 & 0.46 & 55    & 98   & 93 \\ \hline
MHT\_DAM-i  & \bf Town    & 0.56 (0.45)  & 0.83 & 0.42 & 0.40 & 0.90 & 0.46 & 59    & 90   & 97 \\ \hline
MHT\_DAM-o  & \bf Town    & 0.57 (0.45)  & 0.81 & 0.44 & 0.42 & 0.89 & 0.48 & 63    & 94   & 89 \\ \hline \hline
NOMT        & \bf Town    & 0.71 (0.62)  & 0.83 & 0.63 & 0.65 & 0.94 & 0.71 & 122   & 76   & 48 \\ \hline
NOMT-i      & \bf Town    & 0.76 (0.65)  & 0.87 & 0.68 & 0.66 & 0.93 & 0.72 & 135   & 61   & 50 \\ \hline
NOMT-o      & \bf Town    & 0.75 (0.63)  & 0.83 & 0.68 & 0.66 & 0.91 & 0.75 & 144   & 59   & 43 \\ \hline \hline
SCEA        & \bf Town    & 0.56 (0.43)  & 0.83 & 0.42 & 0.40 & 0.90 & 0.46 & 56    & 95   & 95 \\ \hline
SCEA-i      & \bf Town    & 0.58 (0.45)  & 0.87 & 0.44 & 0.44 & 0.95 & 0.47 & 62    & 89   & 95 \\ \hline
SCEA-o      & \bf Town    & 0.58 (0.43)  & 0.80 & 0.45 & 0.43 & 0.89 & 0.50 & 65    & 94   & 87 \\ \hline \hline
TDAM        & \bf Town    & 0.60 (0.48)  & 0.71 & 0.52 & 0.39 & 0.78 & 0.56 & 70    & 112  & 64 \\ \hline
TDAM-i      & \bf Town    & 0.60 (0.48)  & 0.73 & 0.51 & 0.41 & 0.80 & 0.56 & 69    & 110  & 67 \\ \hline
TDAM-o      & \bf Town    & 0.59 (0.45)  & 0.67 & 0.54 & 0.37 & 0.74 & 0.60 & 82    & 108  & 56 \\ \hline \hline
TSML\_CDE   & \bf Town    & 0.68 (0.58)  & 0.75 & 0.63 & 0.72 & 0.95 & 0.79 & 143   & 79   & 24 \\ \hline
TSML\_CDE-i & \bf Town    & 0.76 (0.62)  & 0.84 & 0.70 & 0.73 & 0.95 & 0.79 & 150   & 68   & 28 \\ \hline
TSML\_CDE-o & \bf Town    & 0.78 (0.62)  & 0.82 & 0.74 & 0.74 & 0.92 & 0.83 & 161   & 68   & 17 \\ \hline \hline
CNNTCM      & \bf Town    & 0.58 (0.46)  & 0.79 & 0.46 & 0.45 & 0.90 & 0.53 & 63    & 110  & 73 \\ \hline
CNNTCM-i    & \bf Town    & 0.61 (0.46)  & 0.80 & 0.49 & 0.48 & 0.90 & 0.55 & 73    & 96   & 77 \\ \hline
CNNTCM-o    & \bf Town    & 0.62 (0.46)  & 0.77 & 0.52 & 0.48 & 0.87 & 0.59 & 85    & 95   & 66 \\ \hline \hline
KSP         & \bf Town    & 0.41 (0.26)  & 0.47 & 0.36 & 0.64 & 0.93 & 0.73 & 107   & 105  & 34 \\ \hline
KSP-i       & \bf Town    & 0.69 (0.42)  & 0.78 & 0.61 & 0.65 & 0.93 & 0.73 & 118   & 91   & 37 \\ \hline
KSP-o       & \bf Town    & 0.69 (0.42)  & 0.76 & 0.63 & 0.64 & 0.91 & 0.75 & 122   & 88   & 36 \\ \hline \hline
MDP         & \bf Town    & 0.59 (0.45)  & 0.65 & 0.55 & 0.50 & 0.81 & 0.68 & 103   & 97   & 46 \\ \hline
MDP-i       & \bf Town    & 0.66 (0.49)  & 0.72 & 0.61 & 0.54 & 0.83 & 0.71 & 116   & 82   & 48 \\ \hline
MDP-o       & \bf Town    & 0.63 (0.45)  & 0.66 & 0.61 & 0.50 & 0.79 & 0.73 & 113   & 94   & 39 \\ \hline \hline
RNN         & \bf Town    & 0.48 (0.30)  & 0.52 & 0.45 & 0.60 & 0.88 & 0.77 & 122   & 103  & 21 \\ \hline
RNN-i       & \bf Town    & 0.59 (0.36)  & 0.65 & 0.55 & 0.61 & 0.90 & 0.76 & 125   & 98   & 23 \\ \hline
RNN-o       & \bf Town    & 0.53 (0.34)  & 0.57 & 0.50 & 0.59 & 0.89 & 0.77 & 125   & 99   & 22 \\ \hline \hline
SORT        & \bf Town    & 0.62 (0.46)  & 0.81 & 0.50 & 0.57 & 0.98 & 0.61 & 49    & 152  & 45 \\ \hline
SORT-i      & \bf Town    & 0.72 (0.47)  & 0.85 & 0.62 & 0.64 & 0.95 & 0.69 & 96    & 109  & 41 \\ \hline
SORT-o      & \bf Town    & 0.65 (0.46)  & 0.83 & 0.60 & 0.60 & 0.90 & 0.65 & 174   & 58   & 14 \\ \hline
\end{tabular}
\caption{Full results for all methods on the {\bf Town} dataset, when using our
detections  as input  and using  the  results of  state-of-the-art trackers  as
input. Number in brackets in IDF\_1 column indicates result for the distance of
1 m.}
\label{tab:res1}
\end{table}

\begin{table}
  \setlength\tabcolsep{10pt}
\begin{tabular}{|l|l|c|c|c|c|c|c|c|c|c|}
  \hline
  Method    & Dataset     & IDF\_1 & IDPR & IDRC & MOTA & PR   & RC   & MT    & PT   & ML \\ \hline
KSP         & \bf ETH     & 0.45 (0.15)  & 0.45 & 0.45 & 0.47 & 0.72 & 0.71 & 182   & 148  & 22 \\ \hline
KSP-i       & \bf ETH     & 0.62 (0.18)  & 0.71 & 0.54 & 0.48 & 0.75 & 0.57 & 134   & 144  & 74 \\ \hline
KSP-o       & \bf ETH     & 0.57 (0.18)  & 0.59 & 0.67 & 0.49 & 0.67 & 0.76 & 217   & 121  & 14 \\ \hline \hline
MDP         & \bf ETH     & 0.55 (0.20)  & 0.63 & 0.48 & 0.40 & 0.79 & 0.60 & 113   & 194  & 45 \\ \hline
MDP-i       & \bf ETH     & 0.58 (0.21)  & 0.76 & 0.46 & 0.41 & 0.83 & 0.50 & 105   & 143  & 104\\ \hline
MDP-o       & \bf ETH     & 0.58 (0.21)  & 0.64 & 0.62 & 0.41 & 0.72 & 0.69 & 157   & 146  & 49 \\ \hline \hline
RNN         & \bf ETH     & 0.51 (0.21)  & 0.54 & 0.49 & 0.48 & 0.80 & 0.73 & 170   & 162  & 20 \\ \hline
RNN-i       & \bf ETH     & 0.54 (0.21)  & 0.76 & 0.39 & 0.48 & 0.85 & 0.44 & 68    & 184  & 100 \\ \hline
RNN-o       & \bf ETH     & 0.54 (0.21)  & 0.40 & 0.47 & 0.47 & 0.64 & 0.76 & 205   & 127  & 20 \\ \hline \hline
SORT        & \bf ETH     & 0.67 (0.29)  & 0.82 & 0.57 & 0.50 & 0.87 & 0.61 & 130   & 175  & 47 \\ \hline
SORT-i      & \bf ETH     & 0.66 (0.26)  & 0.84 & 0.55 & 0.49 & 0.86 & 0.56 & 136   & 129  & 87 \\ \hline
SORT-o      & \bf ETH     & 0.67 (0.29)  & 0.79 & 0.68 & 0.49 & 0.80 & 0.70 & 167   & 148  & 37 \\ \hline \hline
KSP         & \bf Hotel   & 0.44 (0.14)  & 0.33 & 0.65 & 0.32 & 0.48 & 0.94 & 270   & 40   & 6 \\ \hline
KSP-i       & \bf Hotel   & 0.53 (0.17)  & 0.38 & 0.75 & 0.33 & 0.47 & 0.94 & 273   & 35   & 8 \\ \hline
KSP-o       & \bf Hotel   & 0.53 (0.17)  & 0.38 & 0.77 & 0.30 & 0.46 & 0.94 & 276   & 32   & 8 \\ \hline \hline
MDP         & \bf Hotel   & 0.40 (0.12)  & 0.34 & 0.46 & 0.33 & 0.47 & 0.64 & 133   & 92   & 91 \\ \hline
MDP-i       & \bf Hotel   & 0.50 (0.13)  & 0.43 & 0.37 & 0.38 & 0.60 & 0.52 & 83    & 110  & 123 \\ \hline
MDP-o       & \bf Hotel   & 0.37 (0.10)  & 0.28 & 0.47 & 0.30 & 0.40 & 0.67 & 143   & 105  & 68 \\ \hline \hline
RNN         & \bf Hotel   & 0.40 (0.14)  & 0.30 & 0.58 & 0.39 & 0.46 & 0.90 & 252   & 45   & 19 \\ \hline
RNN-i       & \bf Hotel   & 0.40 (0.14)  & 0.30 & 0.59 & 0.39 & 0.46 & 0.90 & 258   & 38   & 20 \\ \hline
RNN-o       & \bf Hotel   & 0.39 (0.13)  & 0.29 & 0.56 & 0.38 & 0.46 & 0.90 & 256   & 41   & 19 \\ \hline \hline
SORT        & \bf Hotel   & 0.54 (0.20)  & 0.45 & 0.68 & 0.37 & 0.55 & 0.82 & 207   & 87   & 22 \\ \hline
SORT-i      & \bf Hotel   & 0.60 (0.20)  & 0.46 & 0.78 & 0.47 & 0.52 & 0.90 & 240   & 60   & 16 \\ \hline
SORT-o      & \bf Hotel   & 0.58 (0.20)  & 0.46 & 0.78 & 0.35 & 0.53 & 0.88 & 238   & 64   & 14 \\ \hline \hline
KSP         & \bf Station & 0.32   & 0.27 & 0.40 & 0.23 & 0.61 & 0.90 & 10166 & 1985 & 211 \\ \hline
KSP-i       & \bf Station & 0.42   & 0.35 & 0.52 & 0.19 & 0.60 & 0.91 & 10296 & 1879 & 187 \\ \hline
KSP-o       & \bf Station & 0.40   & 0.32 & 0.53 & 2.27 & 0.55 & 0.92 & 10597 & 1576 & 189 \\ \hline \hline
MDP         & \bf Station & 0.48   & 0.39 & 0.63 & 0.51 & 0.56 & 0.90 & 9362  & 2293 & 437 \\ \hline
MDP-i       & \bf Station & 0.47   & 0.36 & 0.65 & 0.52 & 0.51 & 0.92 & 10047 & 1771 & 544 \\ \hline
MDP-o       & \bf Station & 0.47   & 0.37 & 0.66 & 0.50 & 0.52 & 0.92 & 10010 & 1930 & 422 \\ \hline \hline
RNN         & \bf Station & 0.30   & 0.24 & 0.37 & 0.40 & 0.58 & 0.90 & 9826  & 2333 & 203 \\ \hline
RNN-i       & \bf Station & 0.30   & 0.24 & 0.38 & 0.41 & 0.59 & 0.90 & 9900  & 2260 & 202 \\ \hline
RNN-o       & \bf Station & 0.30   & 0.25 & 0.39 & 0.40 & 0.57 & 0.90 & 9898  & 2265 & 199 \\ \hline \hline
SORT        & \bf Station & 0.50   & 0.50 & 0.50 & 0.32 & 0.71 & 0.72 & 5557  & 6181 & 624 \\ \hline
SORT-i      & \bf Station & 0.50   & 0.47 & 0.54 & 0.31 & 0.69 & 0.78 & 6996  & 4882 & 484 \\ \hline
SORT-o      & \bf Station & 0.52   & 0.48 & 0.57 & 0.31 & 0.67 & 0.79 & 7154  & 4703 & 505 \\ \hline
\end{tabular}
\caption{Full results  for all methods on  all the datasets except  {\bf Town},
when using  our detections as input  and using the results  of state-of-the-art
trackers as input. Number in brackets in IDF\_1 column indicates result for the
distance of 1 m.}

\label{tab:res2}
\end{table}

\begin{table}
\setlength\tabcolsep{10pt}
\begin{tabular}{|l|l|c|c|c|c|c|c|c|c|c|}
\hline
Method & Dataset     & IDF\_1          & IDPR & IDRC & MOTA     & PR   & RC   & MT    & PT   & ML \\ \hline
KSP    & \bf Town    & 0.56 (0.47)     & 0.55 & 0.57 & 0.87     & 0.93 & 0.97 & 226   & 8    & 12 \\ \hline
MDP    & \bf Town    & 0.87 (0.84)     & 0.92 & 0.82 & 0.87     & 0.99 & 0.89 & 184   & 38   & 24 \\ \hline
RNN    & \bf Town    & 0.65 (0.57)     & 0.65 & 0.65 & 0.85     & 0.95 & 0.95 & 222   & 19   & 5 \\ \hline
SORT   & \bf Town    & 0.88 (0.85)     & 0.93 & 0.84 & 0.90     & 1.00 & 0.90 & 203   & 34   & 9 \\ \hline
OUR    & \bf Town    & \bf 0.97 (0.92) & 0.97 & 0.97 & \bf0.98  & 1.00 & 1.00 & 245   & 1    & 0 \\ \hline \hline
KSP    & \bf ETH     & 0.59 (0.12)     & 0.58 & 0.60 & 0.70     & 0.87 & 0.89 & 287   & 56   & 9 \\ \hline
MDP    & \bf ETH     & 0.89 (0.18)     & 0.91 & 0.87 & 0.85     & 0.95 & 0.91 & 300   & 42   & 10 \\ \hline
RNN    & \bf ETH     & 0.65 (0.16)     & 0.64 & 0.65 & 0.73     & 0.89 & 0.90 & 289   & 62   & 1 \\ \hline
SORT   & \bf ETH     & \bf 0.93 (0.20) & 0.98 & 0.88 & 0.85     & 0.97 & 0.87 & 307   & 31   & 14 \\ \hline
OUR    & \bf ETH     & 0.92 (0.19)     & 0.92 & 0.92 & \bf 0.94 & 0.98 & 0.98 & 347   & 5    & 0 \\ \hline \hline
KSP    & \bf Hotel   & 0.60 (0.21)     & 0.61 & 0.58 & 0.74     & 0.90 & 0.86 & 217   & 69   & 30 \\ \hline
MDP    & \bf Hotel   & 0.85 (0.33)     & 0.87 & 0.83 & 0.84     & 0.95 & 0.90 & 249   & 37   & 30 \\ \hline
RNN    & \bf Hotel   & 0.70 (0.28)     & 0.69 & 0.71 & 0.78     & 0.91 & 0.94 & 284   & 29   & 3 \\ \hline
SORT   & \bf Hotel   & 0.88 (0.36)     & 0.97 & 0.81 & 0.82     & 0.99 & 0.83 & 191   & 107  & 18 \\ \hline
OUR    & \bf Hotel   & \bf 0.94 (0.38) & 0.94 & 0.94 & \bf 0.97 & 1.00 & 1.00 & 314   & 1    & 1 \\ \hline \hline
KSP    & \bf Station & 0.45            & 0.44 & 0.45 & \bf 0.80 & 0.93 & 0.95 & 10957 & 832  & 573 \\ \hline
MDP    & \bf Station & 0.75            & 0.70 & 0.80 & 0.68     & 0.81 & 0.93 & 464   & 67   & 51 \\ \hline
RNN    & \bf Station & 0.40            & 0.39 & 0.40 & 0.68     & 0.90 & 0.94 & 10870 & 1244 & 248 \\ \hline
SORT   & \bf Station & \bf 0.72        & 0.85 & 0.63 & 0.70     & 1.00 & 0.74 & 4968  & 6481 & 913 \\ \hline
OUR    & \bf Station & 0.70            & 0.62 & 0.62 & 0.77     & 0.99 & 0.99 & 579   & 3    & 0 \\ \hline

\end{tabular}
\caption{Full results for all combinations  of methods and datasets, when using
our  set of  ground  truth  detections. Number  in  brackets  in IDF\_1  column
indicates result for the distance of 1 m.}
\label{tab:res3}
\end{table}

\clearpage

\section{Running time evaluation}
\label{sec:time}
\begin{figure*}[!h]
\begin{center}
\hspace{-1pt}
\begin{tabular}{ccc}
  \hspace{-0.4cm}\includegraphics[width=0.33\textwidth]{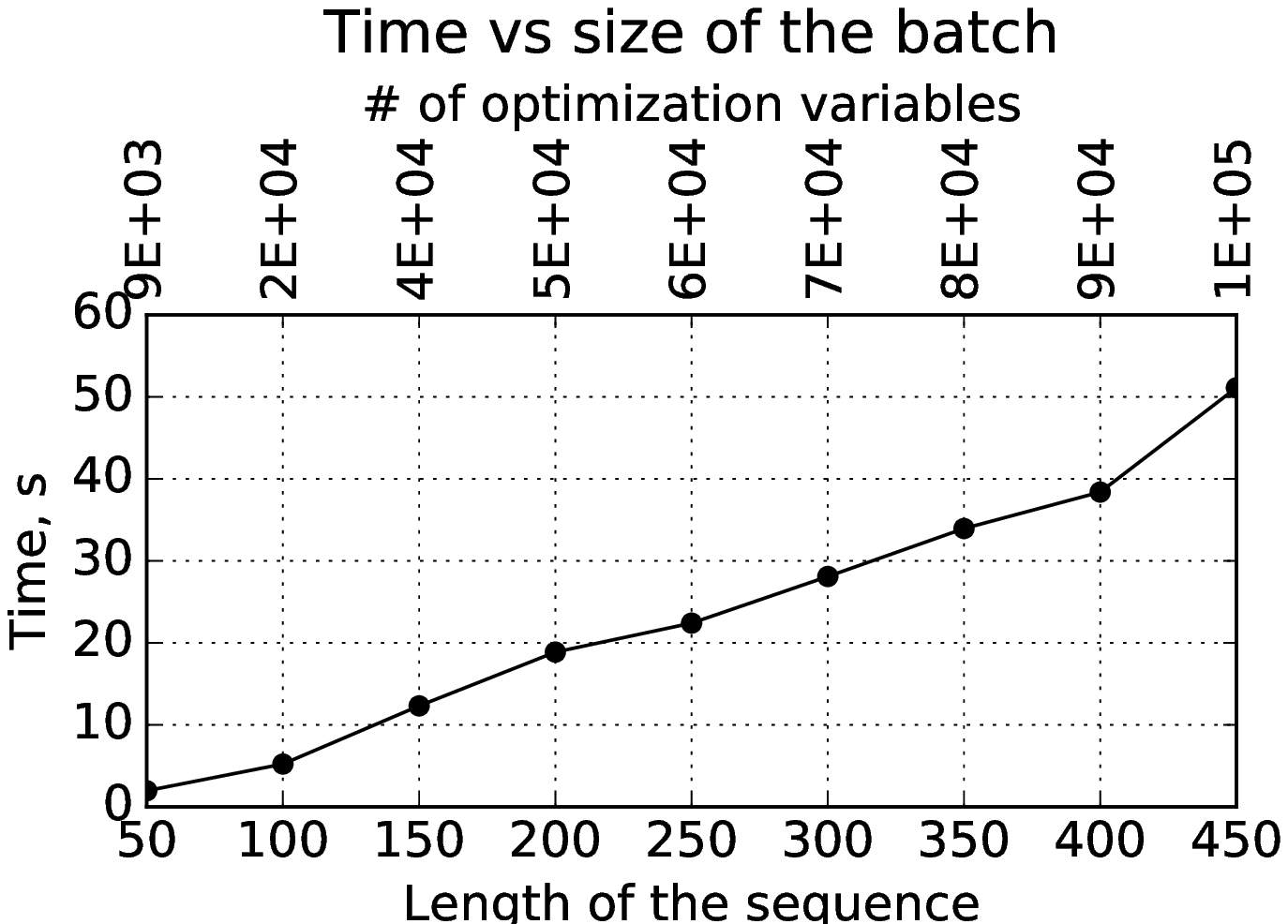} &
  \hspace{-0.4cm}\includegraphics[width=0.33\textwidth]{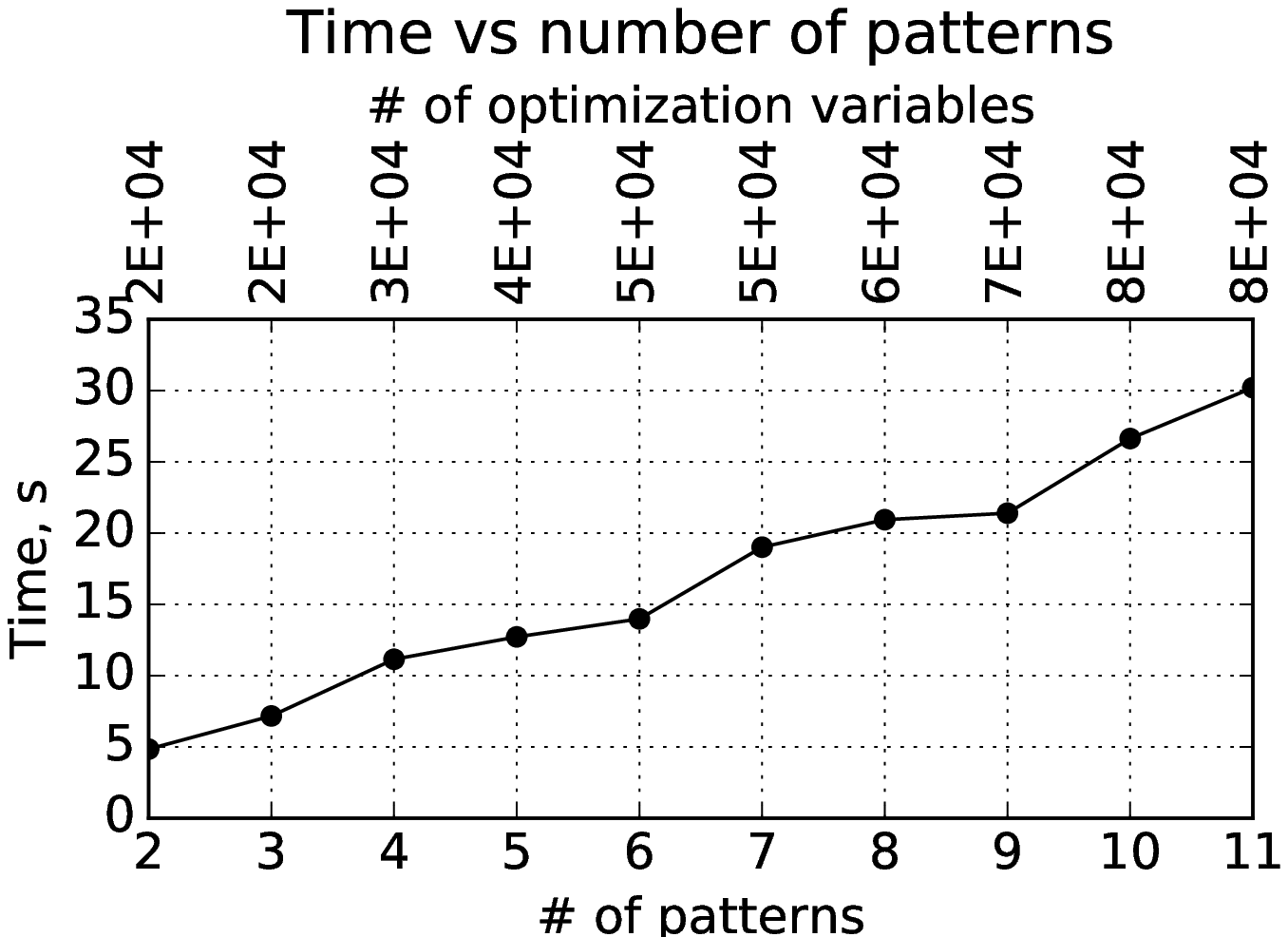} &
  \hspace{-0.4cm}\includegraphics[width=0.33\textwidth]{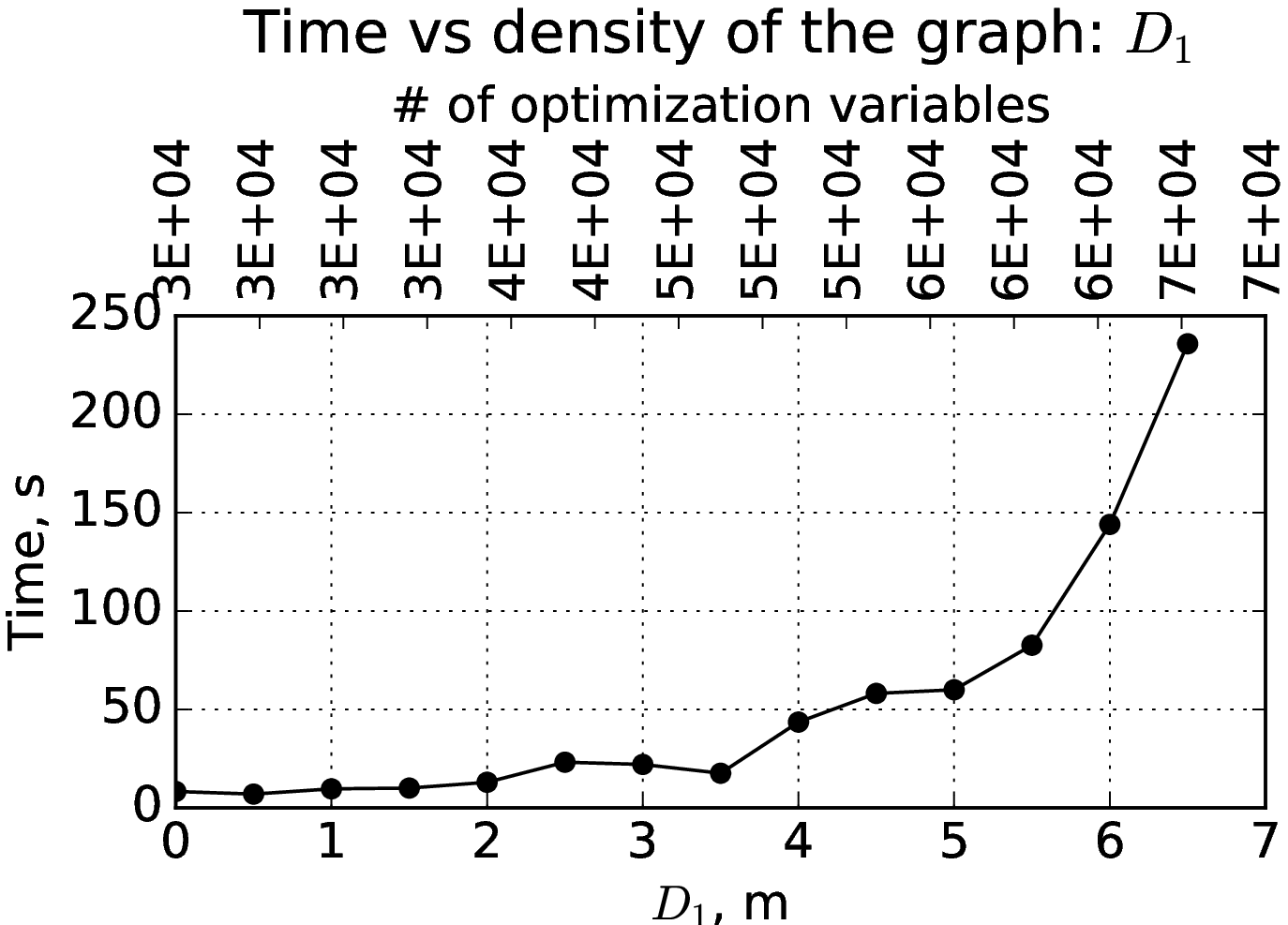} \\
   \hspace{-0.2cm} (a) &
   \hspace{-0.2cm} (b) &
   \hspace{-0.2cm} (c) \\
   \hspace{-0.4cm}\includegraphics[width=0.33\textwidth]{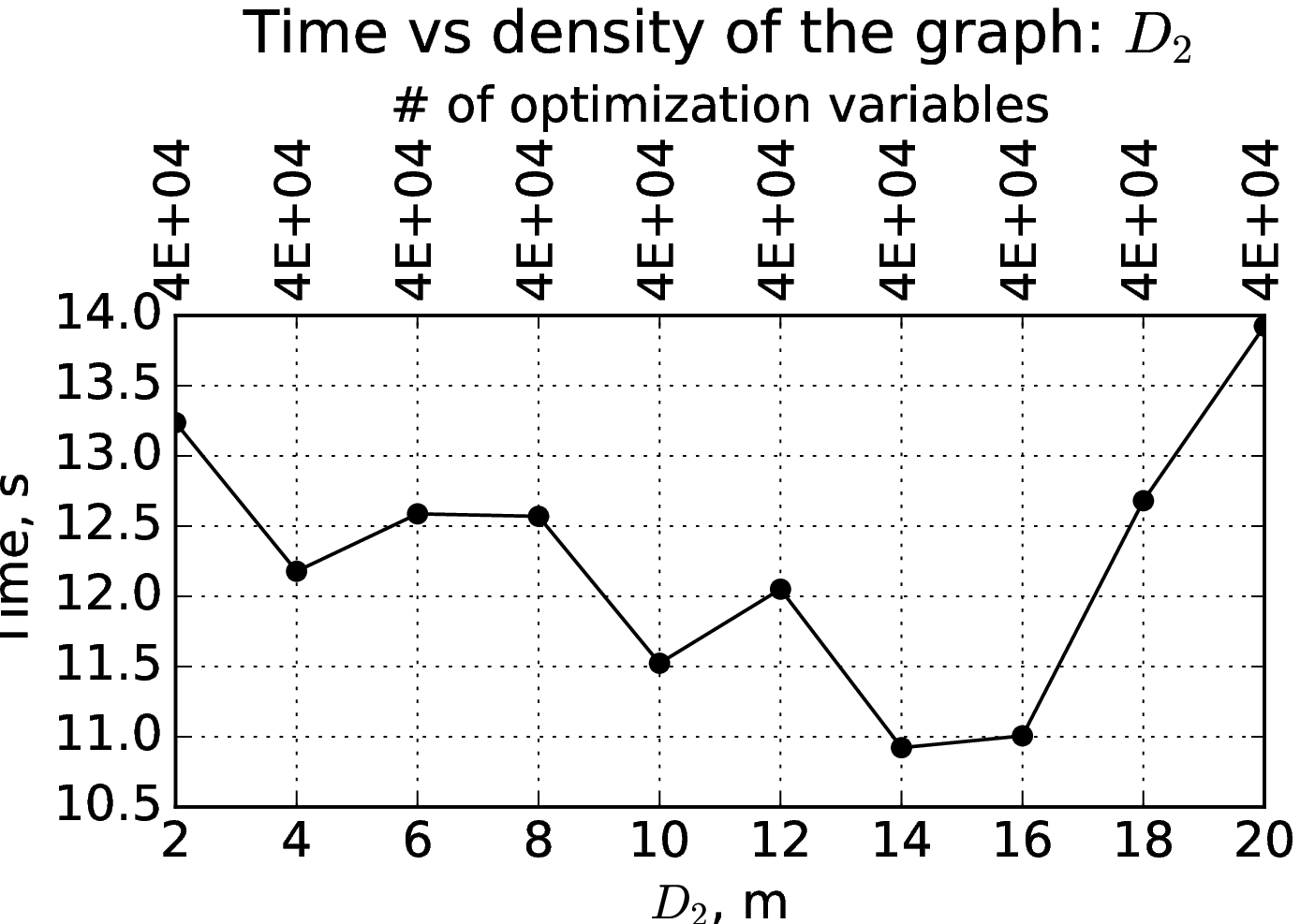}&
  \hspace{-0.4cm}\includegraphics[width=0.33\textwidth]{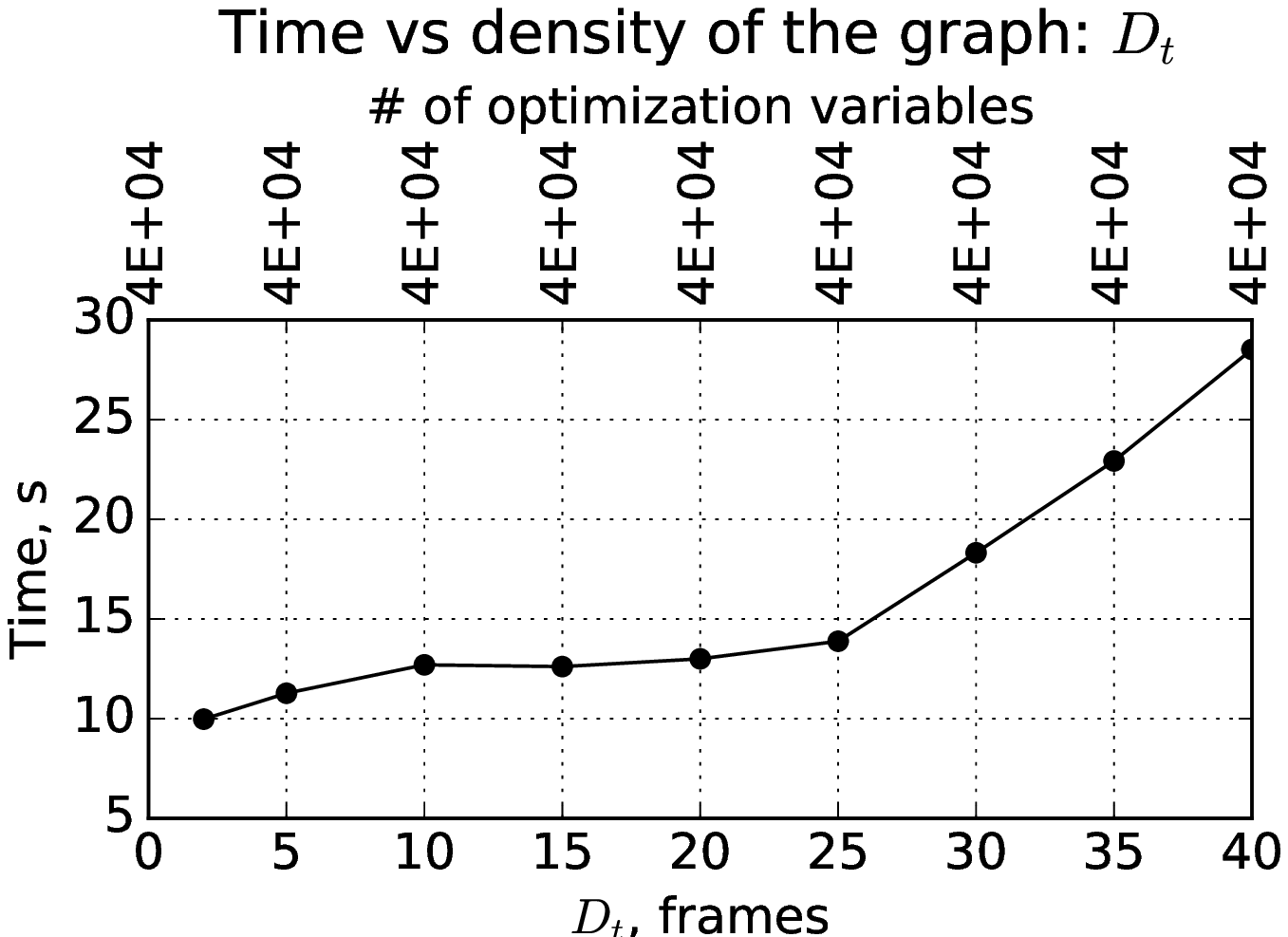} &
  \hspace{-0.4cm}\includegraphics[width=0.33\textwidth]{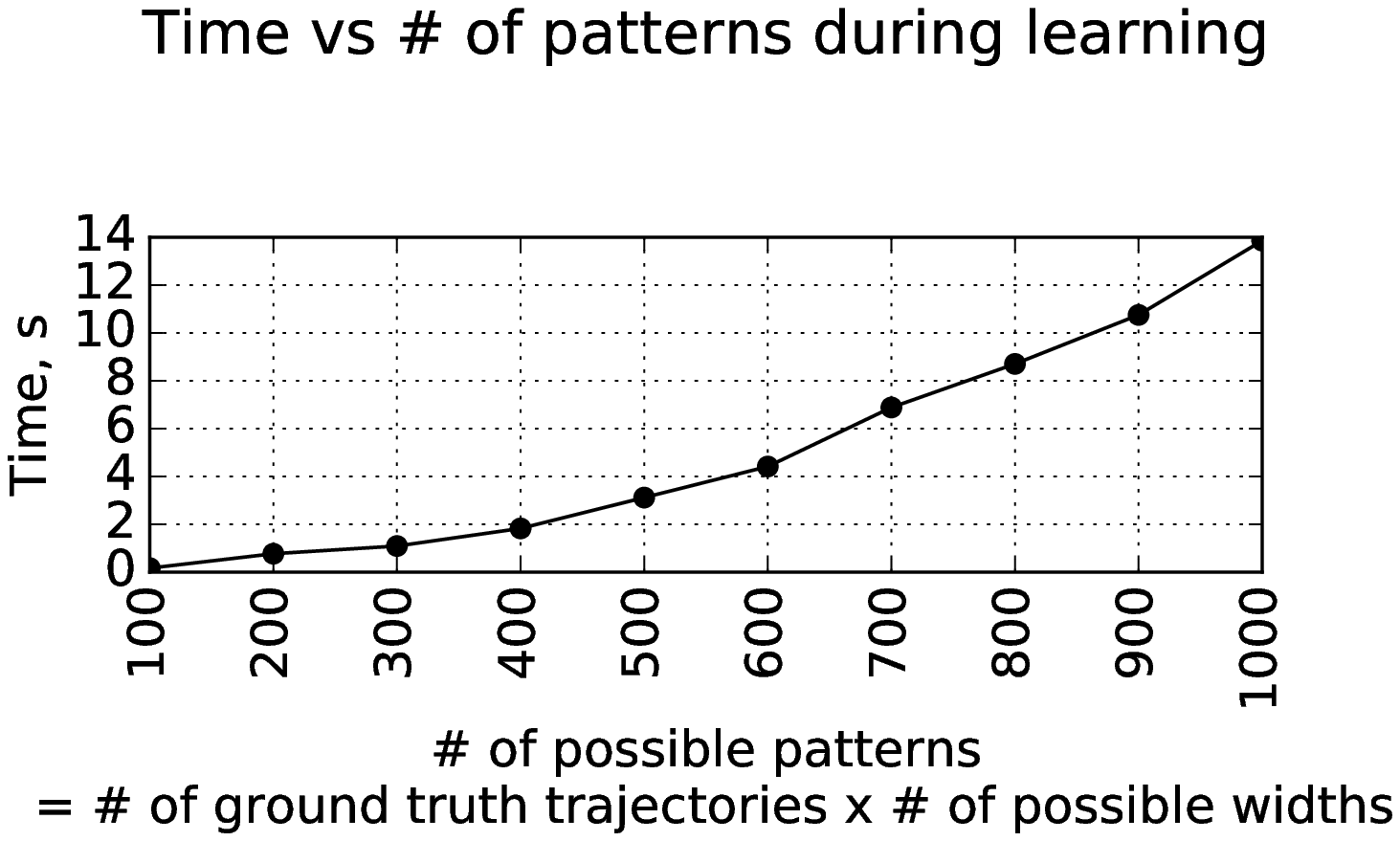} \\
   \hspace{-0.2cm} (d) &
   \hspace{-0.2cm} (e) &
   \hspace{-0.2cm} (f) \\
\end{tabular}
\end{center}
\vspace{-0.4cm}
\caption{The running time and the number of variables of the optimization for tracking are approximately}
\begin{itemize}
\item linear with respect to the number of frames in the batch	{\bf (a)},
\item linear with respect to the number of patterns {\bf(b)},
\item superlinear with respect to the maximum distance at which we join the detections in the neighbouring frames $D_1$, as it directly affects the density of the tracking graph {\bf(c)},
\item almost independent from the maximum distance in space $D_2$ and it time $D_t$ at which we join the endings and beginning of the input trajectories $D_2$, as it has almost no effect on the density of the tracking graph {\bf(d), (e)};
\item The running time and the number of variables of the optimization for learning patterns grows quadraticaly with the number of input trajectories, as each of them is both a trajectory that needs to be assigned to a pattern, and a possible centerline of a pattern {\bf (f)}.
\end{itemize}
\label{fig:time}
\end{figure*}

Here  we present  the  evaluation of  running  time  of  our
approach depending on  the parameters of the optimization. As  mentioned in the
Section 6.4 of the paper and shown in Fig.~\ref{fig:time}, the optimization
time depends  mostly on the  number of  possible transitions between   people,
which  is  controlled by  $D_1$.  The  time  for  learning the  patterns  grows
approximately quadratically.

